%% file: main.tex
\documentclass{IEEEtran}

\input{definition}
\let\csgzli\relax
\begin{document}

\title{\AutoShape: An Autoencoder-Shapelet Approach for Time Series Clustering}

\author{Guozhong~Li,
  Byron~Choi,
  Jianliang~Xu,
  Sourav~S~Bhowmick,\\
  Daphne Ngar-yin Mah,
  and~Grace~Lai-Hung~Wong
  \thanks{Manuscript received XX XX, XXXX; revised XX XX, XXXX.}
\IEEEcompsocitemizethanks{\IEEEcompsocthanksitem Guozhong Li, Byron Choi,
  and Jianliang Xu are with the Department of Computer Science, Hong Kong Baptist University, Hong Kong SAR
(e-mail: \{csgzli, bchoi, xujl\}@comp.hkbu.edu.hk).
\IEEEcompsocthanksitem Sourav~S~Bhowmick is with School of Computing Engineering,
Nanyang Technological University, Singapore (e-mail: assourav@ntu.edu.sg).
\IEEEcompsocthanksitem Daphne Ngar-yin Mah is with Asian Energy Studies Centre and Department of Geography,
Hong Kong Baptist University, Hong Kong SAR (e-mail: daphnemah@hkbu.edu.hk).
\IEEEcompsocthanksitem Grace L.H. Wong is with Medical Data Analytic Centre (MDAC) and Department of Medicine and Therapeutics,
The Chinese University of Hong Kong, Hong Kong SAR (e-mail: wonglaihung@cuhk.edu.hk).}
}


\maketitle

\begin{abstract}
  Time series shapelets are discriminative subsequences that
  have been recently found effective for time series clustering (\TSC).
  The shapelets are convenient for interpreting the clusters.
  Thus, the main challenge for \TSC~is to discover high-quality variable-length shapelets to discriminate different clusters.
  In this paper, we propose a novel {\em autoencoder-shapelet} approach (\AutoShape),
  which is the first study to take the advantage of both autoencoder and shapelet for determining shapelets in an unsupervised manner.
  An autoencoder is specially designed to learn high-quality shapelets.
  More specifically, for guiding the latent representation learning, we employ the latest self-supervised loss
  to learn the unified embeddings for variable-length shapelet candidates (time series subsequences) of different variables,
  and propose the diversity loss to select the discriminating embeddings in the unified space.
  We introduce the reconstruction loss to recover shapelets in the original time series space for clustering.
  Finally, we adopt Davies–Bouldin index (DBI) to inform \AutoShape\ of the clustering performance during learning.
  We present extensive experiments on \AutoShape.
  To evaluate the clustering performance on univariate time series (\UTS),
  we compare \AutoShape\ with $15$ representative methods using UCR archive datasets.
  To study the performance of multivariate time series (\MTS),
  we evaluate \AutoShape\ on $30$ UEA archive datasets with $5$ competitive methods.
  The results validate that \AutoShape\ is the best among all the methods compared.
  We interpret clusters with shapelets, and can obtain interesting intuitions about clusters
  in two \UTS~case studies and one \MTS~case study, respectively.
  \end{abstract}


\input{./introduction}
\input{./related}
\input{./ourmethod}
\input{./exp}
\input{./conclusion}

\eat{\stab
\noindent{\bf Acknowledgements.}
The authors would like to thank the anonymous reviewers for the helpful comments and suggestions.}

\bibliographystyle{abbrv}
\bibliography{main}

\input{./bio}

\end{document}

%% file: definition.tex
\usepackage{xcolor}
\usepackage{array}
\usepackage{subfigure}
\usepackage{graphicx}
\usepackage{url}
\usepackage{hyperref}
\usepackage{amsfonts}
\usepackage{amsmath}
\usepackage[linesnumbered,titlenumbered,ruled,vlined,commentsnumbered,noend]{algorithm2e}

\newtheorem{example}{Example}

\newtheorem{definition}{Definition}

\newcommand{\csgzli}[1]{\textcolor{blue}{#1}}

\newcommand*\numcircledmod[1]{\raisebox{.5pt}{\textcircled{\raisebox{-.9pt} {#1}}}}
\newcommand{\stab}{\rule{0pt}{8pt}\\[-2.0ex]}

\newcommand{\eat}[1]{}
\newcommand{\eg}{{\it e.g.}, }
\newcommand{\ie}{{\it i.e.}, }
\newcommand{\qed}{$\hfill\square$}

\newcommand{\TSC}{{\sc tsc}}
\newcommand{\UTS}{{\sc uts}}

\newcommand{\MTS}{{\sc mts}}

\newcommand{\AutoShape}{{\sc AutoShape}}

\newcommand{\kmeans}{\mathsf{kmeans}}
\newcommand{\dist}{\mathsf{dist}}

\newcommand{\sort}{\mathsf{sort}}

\newcommand{\pop}{\mathsf{pop}}
\newcommand{\append}{\mathsf{append}}

%% file: introduction.tex

\section{Introduction}\label{intro}
\begin{figure}[tbp]
    \centering
    \includegraphics[width=.9\linewidth]{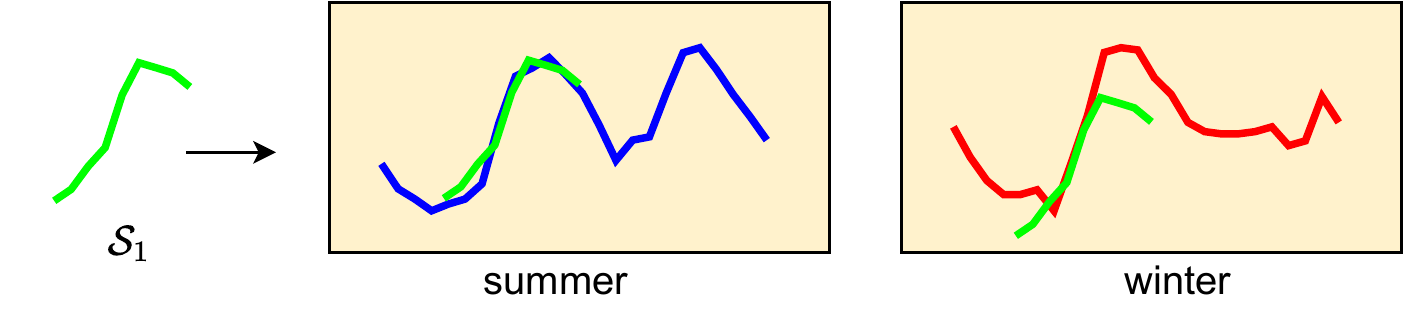}
    \caption{The shapelet $S_1$ for ItalyPowerDemand~\cite{UCRArchive},
      which was derived from daily electrical power consumption time series from Italy in 1997.
      Two classes are in the dataset, summer (blue) from April to September, winter (red) from October to March.}\label{fig:italy-intro}
\end{figure}

\IEEEPARstart{T}{ime} series clustering (\TSC) has numerous applications in both academia and industry~\cite{aghabozorgi2015time,guo2021trend,he2021csmvc},
and thus many research approaches~\cite{asadi2016creating,guo2021information,liao2005clustering} for solving \TSC\ have been proposed.
The classical approaches to solving the \TSC\ problem can be categorized as whole series-based,
feature-based, and model-based~\cite{li2019multivariate, singhal2005clustering, zhou2014model}.
These approaches use the raw time series themselves, feature extraction, and model parameter transformation,
and then apply K-means, DBSCAN, or other clustering algorithms.
A recent trend in \TSC\ is to find some local patterns or features from raw time series data.
Among these approaches, {\em shapelet-based methods} (\eg \cite{ulanova2015scalable,zakaria2012clustering,zhang2018salient})
have repeatedly demonstrated their superior performance on \TSC.
Figure~\ref{fig:italy-intro} shows an example of the shapelet $S_1$ discovered
by \AutoShape~from one UCR \UTS~dataset, \ie ItalyPowerDemand~\cite{UCRArchive}.

In the seminal work on shapelets~\cite{ye2009time}, they are discriminative time series subsequences,
providing interpretable results, although shapelets are initially proposed for the classification problem on time series.
The interpretability from shapelets has also been gauged by the cognitive metrics with respect to human~\cite{li2020survey}.
The unsupervised-shapelet (a.k.a. u-shapelet) is first learned from unlabeled time series data for time series clustering in~\cite{zakaria2012clustering}.
Scalable u-shapelet method~\cite{ulanova2015scalable} has been proposed for concerning the efficiency issue of shapelet discovery in \TSC.
To further improve clustering quality, Zhang et al. introduced an unsupervised model for learning shapelets, called USSL~\cite{zhang2018salient}.
\csgzli{STCN~\cite{ma2020self} is proposed to optimize the feature extraction and self-supervised clustering simultaneously.}

\eat{\begin{example}
     From the learned shapelet $S_1$, we can identify that the power demand in summer is lower than that in winter from 5am to 11pm.
     This is because the heating in the morning used during the winter time
     and air conditioning in summer was still fairly rare in Italy when the data were collected~\cite{keogh2006intelligent}.
   \end{example}}

\begin{figure*}[tbp]
  \centering
  \includegraphics[width=.9\linewidth]{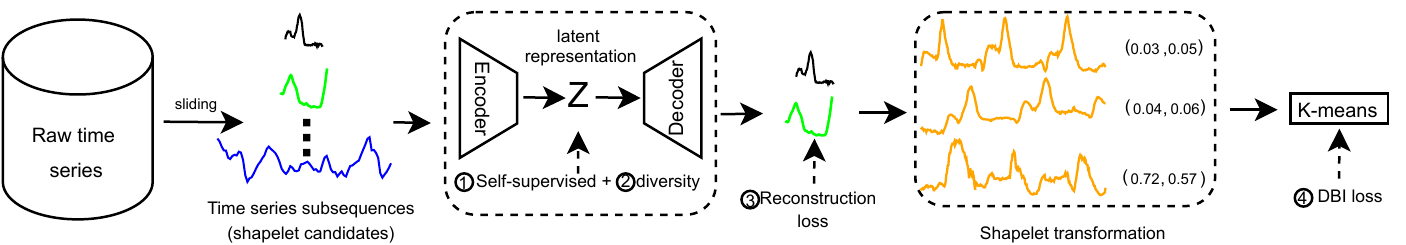}
  \caption{The overview of \AutoShape\ for time series clustering (\TSC).
  After time series subsequences (shapelet candidates) generation using sliding windows,
  \numcircledmod{1}~$\mathcal{L}_{Self-supervised}$ learns latent representations ${\bf Z}$
  for time series subsequences to capture their characteristics.
  \numcircledmod{2}~$\mathcal{L}_{Diversity}$ selects the shapelets with both universality and heterogeneity.
  \numcircledmod{3}~$\mathcal{L}_{Reconstruction}$ makes the reconstruction of the latent representations
  to preserve the shapes of the shapelets.
  Then, the selected shapelets (black and green ones) are exploited
  to transform the original time series (orange color)
  into the transformed representations (using vectors to represent the transformed representations).
  The clustering algorithm (K-means) is applied to the transformed representations.
  \numcircledmod{4}~$\mathcal{L}_{DBI}$ is computed from the clustering results and used to
  adjust the shapelets, to improve the final clustering performance.}\label{fig:Shapelet-Autoencoder}
\end{figure*}

Recently, autoencoder-based methods (\eg DEC~\cite{xie2016unsupervised}, IDEC~\cite{guo2017improved})
have been applied to the clustering problem and produce effective results.
They optimize a clustering objective for learning a mapping from the raw data space to a lower-dimensional space.
However, they are developed for the text and image clustering, not for time series.
Some of autoencoder-based methods (\eg DTC~\cite{madiraju2018deep} and DTCR~\cite{ma2019learning}) are then  proposed for time series.
They utilize an autoencoder network to learn general representations of \textit{whole} time series instances under several different objectives.
The trained encoder network is naturally employed to embed the raw time series into the new representations,
which are used to replace the raw data for final clustering (\eg K-means).
Nonetheless, they focus on whole time series instances, ignore the importance of local features (time series subsequences)
and miss the reasoning of clustering, namely interpretability~\cite{bertsimas2021interpretable}.

Different from the above-mentioned autoencoder-based works, we learn a unified representation
for variable-length time series \textit{subsequences} (namely, \textit{shapelet candidates})
of different variables.
After all the shapelet candidates are embedded into the same latent space,
it becomes simple to measure the similarity among the candidates and to further determine shapelets for clustering.
Importantly, because of the autoencoder approach, shapelets are no longer restricted to real time series subsequences,
which enlarges the scope of shapelet discovery from the raw data~\cite{ye2009time}.

In this paper, we propose a novel autoencoder-based shapelet approach for \TSC\ problem, called \AutoShape.
To the best of our knowledge, this is the first study to take
the advantage of both shapelet-based methods and autoencoder-based methods for \TSC.
An overview of \AutoShape~is presented in Figure~\ref{fig:Shapelet-Autoencoder}.

{\em Four} objectives are specially designed for learning the final shapelets for clustering.
First, the self-supervised loss learns general unified embeddings of time series subsequences (shapelet candidates).
Specifically, we employ a {\em cluster-wise triplet loss} \cite{shapeletnet2020},
which has been shown effective for representing time series.
Second, we propose a {\em diversity loss} to learn top-$k$ candidates after clustering all the embeddings.
The learned candidates, closest to the centroid of a cluster, are of high quality
that have two characteristics: they yield large clusters, and the clusters are distant from each other.
Third, we decode the selected embeddings through the {\em reconstruction loss} to obtain decoded shapelets.
Such shapelets maintain the shapes of the original time series subsequences for human interpretation.
Then, the original time series are transformed with respect to the decoded shapelets.
The new transformed representations are passed to construct a clustering model (\eg K-means).
Fourth, after achieving the clustering result, the {\em Davies–Bouldin index} (DBI)~\cite{davies1979cluster}
is calculated to adjust the shapelets.

An autoencoder network is applied to jointly learn the time series subsequences representations
to select high-quality shapelets for the transformation.
Our shapelets for transformation are decoded by the autoencoder,
not necessarily restricted to the raw time series.
At the same time, the reconstruction loss of \AutoShape~ maintains
the shapes of raw time series subsequences in the final shapelets,
rather than learning some subsequences very different from the original time series.

We conduct comprehensive experiments on both univariate time series (UCR archive)~\cite{UCRArchive}
and multivariate time series (UEA archive)~\cite{UEAArchive}.
The results show that in terms of normalized mutual information (NMI) and rand index (RI)~\cite{zhang2018salient},
\AutoShape\ is the best of all $15$ and $6$ representative methods compared on univariate
and multivariate time series (\UTS~and \MTS), respectively.
We note that \AutoShape\ performs the best in $15$ out of $36$ \UTS~datasets and $24$ out of $30$ \MTS~datasets.
Furthermore, an ablation study verifies the effectiveness of self-supervised loss, diversity loss and DBI objective.
We present three cases from UCR archive~(human motion recognition, power demand, and image)
and one case from UEA archive~(EEG of human activity recognition)
to illustrate the intuitions of the learned shapelets.

The main contributions of this paper are summarized as follows:
\begin{itemize}

\item We propose an autoencoder-based shapelet approach, \AutoShape,
  to jointly learn the time series subsequences representations
  for discovering discriminative shapelets for \TSC~in an unsupervised manner.

\item Four objectives, namely, the self-supervised loss for latent representations,
  diversity loss for both universality and heterogeneity,
  reconstruction loss to preserve the shapes,
  and DBI objective to improve the final clustering performance,
  are specially designed for learning the final shapelets for clustering.

\item Extensive experiments on UCR (\UTS) datasets and UEA (\MTS) datasets for \TSC~verify
  that our \AutoShape~is significantly more competitive in terms of accuracy
  when compared to the state-of-the-art methods.

\item The interpretability of the learned shapelets,
  which may not be the real subsequences from raw time series data,
  is illustrated in four case studies on \UTS~and \MTS~datasets.

\end{itemize}

\noindent{\bf Organization.}
The rest of this paper is organized as follows.
Section~\ref{Related Work} reviews the related work.
The details of our proposed method are given in Section~\ref{autoencoder}.
Section~\ref{experiment} reports the experimental results on both \UTS~and \MTS.
Section~\ref{conclusion} concludes the paper and presents the future work.

%% file: related.tex

\section{Related work}\label{Related Work}
Interested readers may refer to some excellent review papers
on time series clustering~\cite{aghabozorgi2015time,liao2005clustering}.
In this section, we focus on the shapelet-based and autoencoder-based methods.

\subsection{Shapelet-based methods}
The shapelet-based method was introduced with an emphasis on its interpretability in~\cite{ye2009time},
and then studies on logical shapelets~\cite{mueen2011logical},
shapelet transformation~\cite{lines2012shapelet},
learning shapelets~\cite{grabocka2014learning}, Matrix Profile~\cite{yeh2016matrix},
and efficient learning shapelets~\cite{hou2016efficient,bspcover2019} are mainly proposed for time series classification.
The unsupervised-shapelet method (a.k.a u-shapelet) was proposed for time series clustering in~\cite{zakaria2012clustering}.
Scalable u-shapelet method, a hash-based algorithm to discover u-shapelets efficiently, was introduced by Ulanova et al.~\cite{ulanova2015scalable}.
The optimized techniques of SAX~\cite{lin2003symbolic} have been exploited in scalable u-shapelet.
k-Shape~\cite{paparrizos2015k} relies on a scalable iterative refinement procedure to generate homogeneous and well-separated clusters.
A normalized cross-correlation measure to calculate the distance of two time series is employed in k-Shape.
Zhang et al.~\cite{zhang2018salient} proposed an unsupervised salient subsequence learning (USSL) model for \TSC,
which incorporates shapelet learning, shapelet regularization, spectral analysis and pseudo labeling.
USSL is similar to the learning time series shapelets method for classification (\ie LTS~\cite{grabocka2014learning}).
\csgzli{Self-supervised time series clustering network (STCN)~\cite{ma2020self}
  optimizes the feature extraction with one-step time series prediction conducted by RNN
  to capture the temporal dynamics and maintain the local structures of time series.}

Unsupervised-shapelets are discovered without label information,
and thus shapelets can been used for not only time series classification but also time series clustering.
Li et al.~\cite{shapeletnet2020} proposed the ShapeNet framework
to discover shapelets for multivariate time series classification.
In comparison, this paper is the first work to investigate how to discover shapelets
to do the clustering on both univariate and multivariate time series.

\subsection{Autoencoder-based methods}
Deep embedding clustering (DEC)~\cite{xie2016unsupervised} is a general method
that simultaneously learns feature representations and cluster assignments
for many data-driven application domains using deep neural networks.
After learning the low-dimensional feature space, a clustering objective is optimized iteratively.
Guo et al.~\cite{guo2017improved} discovered that the defined clustering loss
may corrupt the feature space and thus cause meaningless representations.
Their proposed algorithm, the improved deep embedding clustering (IDEC),
can preserve the structure of the data generating distribution with an under-complete autoencoder.
Deep temporal clustering (DTC)~\cite{madiraju2018deep} naturally integrates an autoencoder network for dimensionality reduction
and a novel temporal clustering layer for new time series representation clustering into a single end-to-end learning framework without using labels.
DTCR~\cite{ma2019learning} proposes a seq2seq autoencoder representation learning model,
integrating reconstruction task (for autoencoder), K-means task (for hidden representation),
and classification task (to enhance the ability of encoder).
After learning the autoencoder, a classical method (\eg~K-means) is applied to the hidden representation.
As we shall see, this paper specially designs loss functions for an autoencoder to determine shapelets for time series clustering.

%% file: ourmethod.tex

\section{Autoencoder for shapelets (\AutoShape)}\label{autoencoder}
In this section, we propose an autoencoder-based shapelet approach, called \AutoShape.
As the name suggests, \AutoShape~adopts the autoencoder network for shapelet discovery.
It learns the unified embeddings of shapelet candidates and meanwhile
preserves original time series subsequences' shapes to make it possible to understand the intuitions of the clusters.
More specifically, we exploit an autoencoder network to learn the general unified embeddings
of time series subsequences (shapelet candidates) using four objectives,
namely \numcircledmod{1}~self-supervised loss, \numcircledmod{2}~diversity loss,
\numcircledmod{3}~reconstruction loss and \numcircledmod{4}~Davies–Bouldin index (DBI) objective.
\eat{Self-supervised loss is used for learning the embeddings of subsequences.
We employ the diversity loss of shapelet candidates to select diverse candidates for clustering.
A reconstruction loss is used to maintain the interpretability of final shapelets.
DBI is the internal index to guide the network learning for promoting the clustering performance.}
We {\em jointly learn} the shapelets with all the four objectives without labels.
We summarize the notations used and their meanings in Table~\ref{table:summary of notations}.

\begin{table}[tbp]
\centering
\caption{Summary of frequently used notations}\label{table:summary of notations}
\resizebox{.8\linewidth}{!}{
  \begin{tabular}{c|l}
\hline
  Notation & Meaning \\
\hline
\hline
    $T$          & a univariate time series (\UTS) instance \\
                 & $T = (t_1,t_2,\cdots,t_i, \cdots,t_N)$,\\
                 & where $t_i$ is the $i$-th value in $T$ \\
                 & and $N$ is the length of $T$\\
\hline
  $D$            & a \UTS~dataset $(T_1,T_2, \cdots,T_M)$, \\
                 & where $M$ is the number of time series in $D$ \\
\hline
  $T_{a,b}$       & a subsequence $T_{a,b}$ of $T$, $(t_{a}, \cdots,t_{b})$, \\
                 & where $1 \le a \le b \le N$, \\
                 & $a$ and $b$, the beginning and ending positions \\
\hline
                 $\mathcal{C}$  & the label set \\
\hline
  $V$            & the number of variables/observations/dimensions \\
\hline
  $\mathbb{T}$   & a multivariate time series (\MTS) instance \\
                 & $\mathbb{T} = (T^1, T^2,\cdots,T^v\cdots, T^V)$, \\
                 & where $T^v = (t^v_1,t^v_2,\cdots,t^v_i, \cdots,t^v_N)$\\
\hline
  $\mathbb{D}$   & an \MTS\ dataset $(\mathbb{T}_1,\mathbb{T}_2, \cdots,\mathbb{T}_M)$, \\
                 & where $M$ is the number of \MTS\ in $\mathbb{D}$ \\
\hline
  $\mathcal{S}$  & the shapelet set \\
\hline
          \end{tabular}
}
\end{table}

\subsection{Shapelet discovery}\label{shapelet-discovery}
We introduce the self-supervised loss for latent representations,
diversity loss for both universality and heterogeneity,
and reconstruction loss to train an autoencoder in detail.

\subsubsection{Self-supervised loss}\label{cluster-wise}
We aim to learn a unified embedding of variable-length shapelet candidates of different variables.
We adopt a cluster-wise triplet loss~\cite{shapeletnet2020}
as the self-supervised loss to learn the embedding in an unsupervised manner,
since it has been shown effective for representing time series subsequences.
The cluster-wise triplet loss function is defined with
(i) the distance between the anchor and multiple positive samples ($\mathcal{D}_{AP}$),
(ii) that between the anchor and multiple negative samples ($\mathcal{D}_{AN}$),
and (iii) the sum of intra distance $\mathcal{D}_{intra}$ among positives and negatives, respectively.
For self-containedness, we recall the cluster-wise triplet loss below.
\begin{small}
\begin{multline}
  \mathcal{L}_{Self-supervised} = \\ \mathcal{L}_{Triplet}\left(f_e(x), f_e(\pmb{x}^{+}), f_e(\pmb{x}^{-})\right) = \\
              \log \frac{\mathcal{D}_{AP} +\mu}
              {\mathcal{D}_{AN}} + \beta \mathcal{D}_{intra} = \\
              \log \frac{\frac{1}{K^{+}} \sum\limits_{i=1}^{K^{+}} ||f_e(x) - f_e(x^{+}_{i})||_2^2 +\alpha}
                   {\frac{1}{K^{-}} \sum\limits_{i=1}^{K^{-}} ||f_e(x) - f_e(x^{-}_{i})||_2^2 } + \beta \mathcal{D}_{intra}
\end{multline}
\end{small}
where $x$ is the anchor, ${\pmb{x}^{+}}$ and ${\pmb{x}^{-}}$ denote the set of positive and negative samples, respectively.
$\mu$ denotes a margin, and $\beta$ is a hyperparameter.

The distances among the positive (negative) samples are included and should be small (large).
The maximum distance among all positive (negative) samples is presented in Formula~\ref{intra-positive} (Formula~\ref{intra-negative}).
\begin{small}
\begin{equation}\label{intra-positive}
  \mathcal{D}_{pos} = \max_{i, j \in (1, K^+) \wedge i<j}\{ ||f(x_i^{+}) - f(x^{+}_{j})||_2^2 \}
\end{equation}
\end{small}
and
\begin{small}
\begin{equation}\label{intra-negative}
  \mathcal{D}_{neg} = \max_{i, j \in (1, K^-) \wedge i<j}\{ ||f(x_i^{-}) - f(x^{-}_{j})||_2^2 \}
\end{equation}
\end{small}

The intra-sample loss is defined as follows:
\begin{small}
\begin{equation}\label{intra-loss}
  \mathcal{D}_{intra} = \mathcal{D}_{pos} + \mathcal{D}_{neg}
\end{equation}
\end{small}
The detail of a differentiable approximation to the maximum function of
$\mathcal{D}_{pos}$ and $\mathcal{D}_{neg}$ was introduced in~\cite{shapeletnet2020}.
Interested readers may refer to an excellent paper~\cite{boyd2004convex}.

The encoder network $f_e(\cdot)$ maps from an original time series space to a latent space.
The embedding function is $f_e: x \rightarrow h$.
The function is learned using the self-supervised loss.
According to~\cite{shapeletnet2020}, the encoder network can be parameterized by any neural network architecture of choice,
with the only requirement that they obey causal ordering (\ie no future value impacts the current value).
Here, we implement the encoder network using a Temporal Convolutional Network (TCN)~\cite{bai2018empirical}.
We also implement a recurrent network, namely vanilla RNN~\cite{yoon2019time}, for the autoencoder.
The network comparison experiment, illustrated in the supplementary material (Section~\ref{network-comparision}),
shows the similar results from both TCN and RNN.
Thus, we use TCN as the default network in the following experiments.

\subsubsection{Diversity loss}
We propose our diversity loss for the autoencoder to discover diverse shapelets of high quality.

Following the protocol of selecting shapelets with diversity
in previous research, namely USSL~\cite{zhang2018salient} and DTCR~\cite{ma2019learning},
we select the diverse shapelets for shapelet transformation in Section~\ref{clustering-mts-st}.
We cluster the shapelet candidates in the new representation space.
After clustering, several clusters of representations are generated.
We select the candidate, which is the closest to each centroid in each cluster.
We propose the diversity loss (Formula~\ref{loss-diversity}),
which considers both (i) the size of each cluster and (ii)
the distances among all the selected candidates to achieve discrimination.
\begin{equation}\label{loss-diversity}
  \mathcal{L}_{Diversity} = e^{-\sum\limits_{i=1}^Y ( \log(i.size) + \log\sum\limits_{j=1}^Y ||f_e(x_i) - f_e(x_j)||_2^2 )}
\end{equation}
where $f_e(x_i)$ is the closest representation to the centroid of cluster $i$,
$f_e(\cdot)$ is the encoder network, and $Y$ is the number of cluster.

The first part of the exponent denotes the size of the $i$th cluster.
The second part is the distance between the representation of the $i$th cluster
and the representations of other clusters.
The diversity loss is designed for selecting shapelets with two characteristics.
The cluster size of the representation determines the universality of the candidate
and the distance shows the heterogeneity of the clusters.

\subsubsection{Reconstruction loss}
We next present a decoder network $f_d(\cdot)$,
which is guided by Mean Square Error (MSE)~\cite{lehmann2006theory} as the reconstruction loss.
The representation $h = f_e(x)$ is the input for the decoder
to reconstruct the original subsequence under MSE loss.
\begin{equation}\label{loss-mse}
  \mathcal{L}_{Reconstruction} = ||x - \tilde x||_2^2
\end{equation}
where the recovery function is $f_d: h \rightarrow \tilde x$ and $\tilde x$ is the decoded time series subsequence.

\noindent
{\bf Analysis.}
The traditional triplet loss~\cite{schroff2015facenet}
only considers one anchor, one positive and one negative,
which do not employ sufficient contextual insight of neighborhood structure
and the triplet terms are not necessarily consistent~\cite{rippel2016metric}.
For learning the general embedding of input data, we propose the self-supervised loss,
which considers many positives and many negatives to penalize.
Our diversity loss further considers two aspects (the size for universality and the distance for diversity)
for selecting shapelets with high quality for shapelet transformation.
Reconstruction loss supports the interpretability of final shapelets.

\subsection{Shapelet adjustment}

After shapelet discovery, we use shapelets to transform
the original time series into a transformed representation,
where each representation is a vector and each element is the Euclidean distance between
the original time series and one of the shapelets.

Specifically, shapelet transformation~\cite{lines2012shapelet} is a method
to transform a time series $T_j$, w.r.t. the shapelets $\mathcal{S}:\{S_1,\cdots,S_k\}$,
into a new data space ($d_1^j$,$\cdots$,$d_k^j$), where $d_i^j$ = $\dist(T_j^v, S_i)$,
and $\dist(T_j^v, S_i)$ is the distance between $T_j^v$ and a shapelet $S_i$ in $\mathcal{S}$.
The distance of two subsequences can be calculated by Formula~\ref{equation:subsequence distance}.

\begin{definition} \textbf{Distance between two time series  (subsequences)}~\cite{fang2018efficient}.
    The distance of the sequence $T_p$ of the length $|T_p|$ and $T_q$ of the length $|T_q|$
    is denoted as (w.l.o.g.\ assuming $|T_q| \ge |T_p|$),
    \begin{small}
      \begin{multline}\label{equation:subsequence distance}
        {\dist(T_p,T_q)} = \min\limits_{j=1, \cdots,|T_q|-|T_p|+1} \frac{1}{|T_p|}
        \sum_{l=1}^{|T_p|} {(tq_{j+l-1}-tp_{l})}^2
      \end{multline}
    \end{small}
    where $tq_i$ and $tp_i$ are the $i$-th value of $T_p$ and $T_q$, respectively. \qed
\end{definition}

Intuitively, $\dist$ calculates the distance of the shorter sequence $T_p$
to the most similar subsequence in $T_q$ (namely, best match location).

\subsubsection{DBI loss}
We apply a classical clustering method (\eg~K-means) on the transformed representations
and then propose a DBI objective to inform some adjustments of the shapelets.
DBI is chosen because it does not need the ground truth for measurement,
which is consistent with the unsupervised learning of \AutoShape.
\begin{small}
\begin{equation}\label{loss-dbi}
  \mathcal{L}_{DBI} = \frac{1}{C_{num}}\sum_{i=1}^{C_{num}} \max_{j \ne i} \frac{A_i + A_j}{M_{i,j}}
\end{equation}
\end{small}
where $C_{num}$ is the cluster number, $A_i$ is the average distance
between each element of cluster $i$ and the centroid of that cluster, a.k.a., cluster diameter.
$M_{i,j}$ is the distance between cluster centroids $i$ and $j$.

In order to calculate the derivative of the loss function,
all the involved functions of the model need to be differentiable.
However, the maximum function of Formula~\ref{loss-dbi} is not continuous and differentiable.
We, therefore, introduce a differentiable approximation to the maximum function~\cite{boyd2004convex}.
For the sake of organizational clarity, we simplify $R_{i,j} = \frac{A_i + A_j}{M_{i,j}}$.
A differentiable approximation of Formula~\ref{loss-dbi} is stated below.
\begin{small}
\begin{equation}\label{loss-dbi-approx}
  \mathcal{L}_{DBI} \approx \tilde{\mathcal{L}}_{DBI} =
  \frac{1}{C_{num}} \sum_{i=1}^{C_{num}} \frac{\sum\limits_{j=1}^N R_{i,j} \cdot e^{\alpha \cdot R_{i,j}}}
  {\sum\limits_{j=1}^N  e^{\alpha \cdot R_{i,j}}}
\end{equation}
\end{small}
when $\alpha \rightarrow +\infty$, $\tilde{\mathcal{L}}_{DBI}$ approaches the true maximum $\mathcal{L}_{DBI}$.
We found that $\alpha = 50$ is large enough to make Formula~\ref{loss-dbi-approx}
yield exactly the same results as the true maximum.

\subsection{Overall loss function}

Finally, the overall loss $\mathcal{L}_{AS}$ of \AutoShape\ is defined by:
\begin{small}
\begin{multline}\label{loss-overall}
  \mathcal{L}_{AS} = \mathcal{L}_{Reconstruction} + \lambda \mathcal{L}_{Triplet} \\ + \mathcal{L}_{Diversity} + \mathcal{L}_{DBI}
\end{multline}
\end{small}
where $\lambda$ is the regularization parameter.

The overall loss (Formula~\ref{loss-overall}) is minimized to jointly learn the shapelets
for the transformation (as illustrated with Figure~\ref{fig:Shapelet-Autoencoder}).
After shapelet candidate generation,
\numcircledmod{1}~$\mathcal{L}_{Triplet}$ learns latent representations
for shapelet candidates to capture their characteristics.
\numcircledmod{2}~$\mathcal{L}_{Diversity}$ selects the candidates with both universality and heterogeneity.
\numcircledmod{3}~$\mathcal{L}_{Reconstruction}$ makes the reconstruction of the latent representations
to preserve the shapes of the candidates for the interpretability.
Then, the clustering algorithm (\eg K-means) is applied to the representations transformed by the selected shapelet candidates.
\numcircledmod{4}~$\mathcal{L}_{DBI}$ is calculated from the clustering results to adjust the shapelets,
to improve the final clustering performance.
All the loss functions model the encoder network,
but the reconstruction loss and the DBI loss build the decoder network only.

\subsubsection{Overall algorithm}\label{clustering-mts-st}
The overall algorithm for determining final shapelets is presented in
Algorithm~\ref{alg:determining final shapelets and transformation},
which consists of two parts: 1) shapelet discovery process under the autoencoder learning process (Lines 2-13),
and 2) the shapelet transformation process on the raw time series data for \TSC~(Lines 15-21).

\stab \noindent
{\em Shapelet discovery.}
We initialize the final shapelet set (Line 2)
and parameters of the network (Line 3) at the beginning.
Our learning approach iterates for each epoch (Line 4)
and each batch (Line 5) to update the representations and network parameters.
The new representations $\mathcal{H}$ are learned through the encoder network $f_e(\cdot)$ (Line 6)
and then decoded by $f_d(\cdot)$ (Line 7).
We do the clustering on the transformed representation for $\mathcal{L}_{DBI}$ (Line 8).
The overall loss $\mathcal{L}_{AS}$ is calculated (Line 9).
The representations and weights are updated in the negative direction of the stochastic gradient of the loss (Lines 10-11).
We sort all the embeddings of the candidates from $\Omega$ under the $\mathcal{L}_{Diversity}$ (Line 12).
The top-$k$ decoded embeddings of the candidates are selected for final shapelets (Line 13).

\stab \noindent
{\em Shapelet transformation.}
We next explain the shapelet transformation procedure with the \MTS~dataset.
We compute the distance between the shapelet and the time series instance having the same variable (Line 17).
The distance between them is calculated by using the formula at the beginning of Section~\ref{equation:subsequence distance} (Line 18).
After the calculation between one time series instance from the original time series dataset and all the shapelets (Line 16),
the transformed representation of the instance is denoted as $\tilde{\mathbb{T}}_{m}$ (Lines 19-20).
Then, the representation of the original dataset $\mathbb{D}$
is transformed to $\tilde{\mathbb{D}} = \tilde{\mathbb{T}}^{M \times k}$ (Line 21) for the final \TSC.

\begin{algorithm}[t]
  \begin{small}
    \KwIn{\MTS~dataset $\mathbb{D} = \mathbb{T}^{M \times V \times N}$, candidates set $\Omega$,
      learning rate $\eta$, epoch number $\mathcal{E}$, final shapelet number $k$}
    \KwOut{Final shapelets $\mathcal{S}_k$}

    \BlankLine\{Shapelet discovery\} \\
    Initialize $\mathcal{S}$ = $\emptyset$ \;
    Initialize parameters of the autoencoder network $\mathcal{W}$ \;
    \For{epoch $= 1 \rightarrow \mathcal{E}$}{
      \For{$\mathcal{X} \in \Omega$ }{
        $\mathcal{H} \leftarrow f_e(\mathcal{X})$ \tcp*{$\mathcal{L}_{Triplet}, \mathcal{L}_{Diversity}$}
        $\tilde{\mathcal{X}} \leftarrow f_d(\mathcal{H})$ \tcp*{$\mathcal{L}_{Reconstruction}$}
        $\kmeans(ST(\tilde{\mathcal{X}}))$ \tcp*{$\mathcal{L}_{DBI}$}
        $\mathcal{L}_{AS} \leftarrow\mathcal{L}_{AS} \left( \mathcal{X} \right)$ \;
        $\Delta \mathcal{W} \leftarrow -\eta\frac{\partial \mathcal{L}} {\partial \mathcal{W}} $ \;
        $\mathcal{W} \leftarrow \mathcal{W} + \Delta \mathcal{W} $ \;
      }
    }
    $f_e(\Omega).\sort(\mathcal{L}_{Diversity})$ \;
    $\mathcal{S}_k  = f_d(f_e(\Omega)).\pop(k)$ \;

    \BlankLine\{Shapelet transformation\} \\
    \For{$m = \{1,2, \cdots, M\}$}{
      \For{$j = \{1, 2, \cdots, k\}$}{
        $v = S_j.variable$ \;
        $d_{m, j} = {\dist(\mathbb{T}^{v}_m, S_j)}$ \;
        $\tilde{\mathbb{T}}_{m}.\append(d_{m, j})$ \;
      }
      $\tilde{\mathbb{T}}_{m} = <d_{m, 1}, d_{m, 2}, \cdots, d_{m, k}> $ \;
    }
    $\tilde{\mathbb{D}} = \tilde{\mathbb{T}}^{M \times k}$ \;

    \Return~$\mathcal{S}_k$
  \end{small}
 \caption{Determining final shapelets}\label{alg:determining final shapelets and transformation}
\end{algorithm}

\stab \noindent
{\em Complexity analysis.}
After the transformation, the \MTS~dataset $\mathbb{D}$ is reduced from $M \times V \times N$ to $M \times k$,
where $|\mathcal{S}_k|$ = $k$ and $k$ is significantly smaller than $V \times N$.
When the transformation of all the time series instances is done,
some standard clustering methods (\eg~K-means) can be exploited
to do the clustering from the transformed representation.
The time complexity of Lines 2-13 and Lines 15-21 are
$O(\frac{|MVN|}{|\mathcal{X}|} \cdot \mathcal{E} + MVN \cdot \log(MVN) + k)$ and $O(Mk)$, respectively.
Hence, the time complexity of Algorithm~\ref{alg:determining final shapelets and transformation}
is $O(\frac{|MVN|}{|\mathcal{X}|} \cdot \mathcal{E})$.
Therefore, it is not surprising that the network learning part is the bottleneck of the time complexity.

%% file: exp.tex

\section{Experiments}\label{experiment}
In this section, we first present comprehensive experiments conducted \AutoShape~
with $15$ related methods on the UCR (univariate) datasets in Section~\ref{exp-uts}.
We then report the results of \AutoShape~on the UEA (multivariate) datasets
with $5$ related methods particularly in Section~\ref{exp-mts}.
The methods compared with \AutoShape~are the same in STCN~\cite{ma2020self}, DTCR~\cite{ma2019learning},
and USSL~\cite{zhang2018salient}.

\subsection{Experimental Settings}
All the experiments were conducted on a machine
with two Xeon E5-2630v3 @ 2.4GHz (2S/8C) / 128GB RAM / 64 GB SWAP
and two NVIDIA Tesla K80, running on CentOS 7.3 (64-bit).

We follow the default hyperparameters of the network from~\cite{bai2018empirical}.
The following are some important parameters used in our experiment.
The batch size, the number of channels, the kernel size of the convolutional network,
and the network depth are set to $10$, $40$, $3$, and $10$, respectively.
The learning rate is kept fixed at the small value of $\eta = 0.001$,
while the number of epochs for network training is $400$.
$\lambda$ is set to $ 0.01$ for the overall loss function Formula~\ref{loss-overall}.
The number of shapelets is chosen from \{$1, 2, 5, 10, 20$\}.
We try various lengths of sliding windows (namely, the lengths of shapelet candidates)
ranging from \{$0.1, 0.2, 0.3, 0.4, 0.5$\}.
Each number means a ratio of the original time series' length
(\eg $0.1$ means $10\%$ of the original time series' length).
The shapelet number and length follow from LTS~\cite{grabocka2014learning},
ShapeNet ~\cite{shapeletnet2020}, and USSL~\cite{zhang2018salient}.


\subsection{Comparison Methods}\label{baseline}
We compare with $15$ representative \TSC~methods and
give some brief information of each method below.
Interested readers may further refer to the original papers.

 \textbf{K-means}~\cite{hartigan1979algorithm}:
  K-means on the entire original time series.
 \textbf{UDFS}~\cite{yang2011l2}:
  Unsupervised discriminative feature selection with $l_{2,1}$-norm regularized.
 \textbf{NDFS}~\cite{li2012unsupervised}:
  Non-negative discriminative feature selection with non-negative spectral analysis.
 \textbf{RUFS}~\cite{qian2013robust}:
  Robust unsupervised discriminative feature selection with robust orthogonal non-negative matrix factorization.
 \textbf{RSFS}~\cite{shi2014robust}:
  Robust spectral learning for unsupervised feature selection with sparse graph embedding.
 \textbf{KSC}~\cite{yang2011patterns}:
  A pairwise scaling distance for K-means and spectral norm for centroid computation.
 \textbf{KDBA}~\cite{petitjean2011global}:
  Dynamic time warpping barycenter averaging for K-means clustering.
 \textbf{k-Shape}~\cite{paparrizos2015k}:
  A scalable iterative refinement procedure to explore the shapes
  under a normalized cross-correlation measure.
 \textbf{U-shapelet}~\cite{zakaria2012clustering}:
  Discovering shapelets without label for time series clustering.
 \textbf{USSL}~\cite{zhang2018salient}:
  Learning salient subsequences from unlabeled time series
  with shapelet regularization, spectral analysis, and pseudo label.
 \textbf{DTC}~\cite{madiraju2018deep}:
  An autoencoder for temporal dimensionality reduction and a novel temporal clustering layer.
 \textbf{DEC}~\cite{xie2016unsupervised}:
  A method simultaneously learns feature representations and cluster assignments using deep neural networks.
 \textbf{IDEC}~\cite{guo2017improved}:
  Manipulating feature space to scatter data points using a clustering loss as guidance with an autoencoder.
 \textbf{DTCR}~\cite{ma2019learning}:
  Learning cluster-specific hidden temporal representations
  with temporal reconstruction, K-means, and classification.
\csgzli{\textbf{STCN}~\cite{ma2020self}:
  A self-supervised time series clustering framework
  to jointly optimize the feature extraction and time series clustering.}

\subsection{Experiments on Univariate Time Series}\label{exp-uts}
We have followed the protocol used in the previous works, \eg k-Shape~\cite{paparrizos2015k},
USSL~\cite{zhang2018salient}, DTCR~\cite{ma2019learning}, and STCN~\cite{ma2020self}..
Thirty-six datasets from a well-known benchmark of time series datasets, namely UCR archive, were tested.
Detailed information on the datasets can be obtained from~\cite{UCRArchive}.

We take normalized mutual information (NMI)~\cite{zhang2018salient}
as the metric for evaluating the methods.
Since we observe the similar trends in the rand index (RI) results,
we provide the background of RI in the supplementary material (Section 6.2).
The NMI close to 1 indicates high quality clustering~\cite{zakaria2012clustering}.
The results of \AutoShape\ are the mean values of $10$ runs
and the standard deviations of all the datasets are less than $0.005$.

\subsubsection{NMI on univariate time series}\label{sub:exp-overall-accuracy}
All the NMI results of the baselines are taken from the original
papers~\cite{zhang2018salient} and~\cite{ma2020self}, respectively.
The overall NMI results for the 36 UCR datasets are presented in Table~\ref{tab:nmi}.

From Table~\ref{tab:nmi}, we can observe that the overall performance of \AutoShape\
ranks first among the $15$ compared methods.
Moreover, \AutoShape\ performs the best in $10$ datasets,
which is much more than the other methods except STCN.
\csgzli{The 1-to-1 Wins NMI number of \AutoShape\ is at least $1.6$x larger than
1-to-1 Losses of USSL, DTCR, and STCN, and clearly more than that of other methods.}
\AutoShape\ achieves much higher NMI numbers in some datasets,
such as BirdChicken and ToeSegmentation1.
Our results on 1-to-1-Losses datasets are only slightly lower than
those of USSL (\eg~Ham, Lighting2) and DTCR (\eg~Car, ECGFiveDays).

\begin{table*}[t]
  \caption{Normalized mutual information (NMI) comparison on the $36$ datasets from {\sc UCR archive}}\label{tab:nmi}
  \centering
  \resizebox{\linewidth}{!}{
    \begin{tabular}{l|rrrrrrrrrrrrrrrr}
      Dataset  & K-means & UDFS   & NDFS   & RUFS   & RSFS   & KSC    & KDBA   & k-Shape & U-shapelet & DTC  & USSL   & DEC  & IDEC   & DTCR & STCN  & Ours   \\
      \hline
      \hline

Arrow & 0.4816 & 0.524 & 0.4997 & 0.5975 & 0.5104 & 0.524 & 0.4816 & 0.524 & 0.3522 & 0.5 & \textbf{0.6322} & 0.31 & 0.2949 & 0.5513 & 0.524 & 0.5624 \\
Beef & 0.2925 & 0.2718 & 0.3647 & 0.3799 & 0.3597 & 0.3828 & 0.334 & 0.3338 & 0.3413 & 0.2751 & 0.3338 & 0.2463 & 0.2463 & \textbf{0.5473} & 0.5432 & 0.3799 \\
BeetleFly & 0.0073 & 0.0371 & 0.1264 & 0.1919 & 0.2795 & 0.2215 & 0.2783 & 0.3456 & 0.5105 & 0.3456 & 0.531 & 0.0308 & 0.0082 & 0.761 & \textbf{1} & 0.531 \\
BirdChicken & 0.0371 & 0.0371 & 0.3988 & 0.1187 & 0.3002 & 0.3988 & 0.2167 & 0.3456 & 0.2783 & 0.0073 & 0.619 & 0.016 & 0.0082 & 0.531 & \textbf{1} & 0.6352 \\
Car & 0.254 & 0.2319 & 0.2361 & 0.2511 & 0.292 & 0.2719 & 0.2691 & 0.3771 & 0.3655 & 0.1892 & 0.465 & 0.2766 & 0.2972 & 0.5021 & \textbf{0.5701} & 0.497 \\
ChlorineConcentration & 0.0129 & 0.0138 & 0.0075 & 0.0254 & 0.0159 & 0.0147 & 0.0164 & 0 & 0.0135 & 0.0013 & 0.0133 & 0.0009 & 0.0008 & 0.0195 & \textbf{0.076} & 0.0133 \\
Coffee & 0.5246 & 0.6945 & \textbf{1} & 0.2513 & \textbf{1} & \textbf{1} & 0.0778 & \textbf{1} & \textbf{1} & 0.5523 & \textbf{1} & 0.012 & 0.1431 & 0.6277 & \textbf{1} & \textbf{1} \\
DiatomsizeReduction & 0.93 & 0.93 & 0.93 & 0.8734 & 0.8761 & \textbf{1} & 0.93 & \textbf{1} & 0.4849 & 0.6863 & \textbf{1} & 0.803 & 0.514 & 0.9418 & \textbf{1} & \textbf{1} \\
Dist.phal.outl.agegroup & 0.188 & 0.3262 & 0.1943 & 0.2762 & 0.3548 & 0.3331 & 0.4261 & 0.2911 & 0.2577 & 0.3406 & 0.3846 & 0.4405 & 0.44 & 0.4553 & \textbf{0.5037} & 0.44 \\
Dist.phal.outl.correct & 0.0278 & 0.0473 & 0.0567 & 0.1071 & 0.0782 & 0.0261 & 0.0199 & 0.0527 & 0.0063 & 0.0115 & 0.1026 & 0.0011 & 0.015 & 0.118 & \textbf{0.2327} & 0.1333 \\
ECG200 & 0.1403 & 0.1854 & 0.1403 & 0.2668 & 0.2918 & 0.1403 & 0.1886 & 0.3682 & 0.1323 & 0.0918 & 0.3776 & 0.1885 & 0.2225 & 0.3691 & \textbf{0.4316} & 0.3928 \\
ECGFiveDays & 0.0002 & 0.06 & 0.1296 & 0.0352 & 0.176 & 0.0682 & 0.1983 & 0.0002 & 0.1498 & 0.0022 & 0.6502 & 0.0178 & 0.0223 & \textbf{0.8056} & 0.3582 & 0.7835 \\
GunPoint & 0.0126 & 0.022 & 0.0334 & 0.2405 & 0.0152 & 0.0126 & 0.1288 & 0.3653 & 0.3653 & 0.0194 & 0.4878 & 0.002 & 0.0031 & 0.42 & \textbf{0.5537} & 0.4027 \\
Ham & 0.0093 & 0.0389 & 0.0595 & 0.098 & 0.0256 & 0.0595 & 0.0265 & 0.0517 & 0.0619 & 0.1016 & \textbf{0.3411} & 0.1508 & 0.1285 & 0.0989 & 0.2382 & 0.3211 \\
Herring & 0.0013 & 0.0253 & 0.0225 & 0.0518 & 0.0236 & 0.0027 & 0 & 0.0027 & 0.1324 & 0.0143 & 0.1718 & 0.0306 & 0.0207 & \textbf{0.2248} & 0.2002 & 0.2019 \\
Lighting2 & 0.0038 & 0.0047 & 0.0851 & 0.1426 & 0.0326 & 0.1979 & 0.085 & 0.267 & 0.0144 & 0.1435 & \textbf{0.3727} & 0.06 & 0.1248 & 0.2289 & 0.3479 & 0.353 \\
Meat & 0.251 & 0.2832 & 0.2416 & 0.1943 & 0.3016 & 0.2846 & 0.3661 & 0.2254 & 0.2716 & 0.225 & 0.9085 & 0.5176 & 0.225 & \textbf{0.9653} & 0.9393 & 0.9437 \\
Mid.phal.outl.agegroup & 0.0219 & 0.1105 & 0.0416 & 0.1595 & 0.0968 & 0.1061 & 0.1148 & 0.0722 & 0.1491 & 0.139 & 0.278 & 0.2686 & 0.2199 & 0.4661 & \textbf{0.5109} & 0.394 \\
Mid.phal.outl.correct & 0.0024 & 0.0713 & 0.015 & 0.0443 & 0.0321 & 0.0053 & 0.076 & 0.0349 & 0.0253 & 0.0079 & 0.2503 & 0.1005 & 0.0083 & 0.115 & 0.0921 & \textbf{0.2873} \\
Mid.phal.TW & 0.4134 & 0.4276 & 0.4149 & 0.5366 & 0.4219 & 0.4486 & 0.4497 & 0.5229 & 0.4065 & 0.1156 & 0.9202 & 0.4509 & 0.3444 & 0.5503 & 0.6169 & \textbf{0.945} \\
MoteStrain & 0.0551 & 0.1187 & 0.1919 & 0.1264 & 0.2373 & 0.3002 & 0.097 & 0.2215 & 0.0082 & 0.0094 & \textbf{0.531} & 0.3867 & 0.3821 & 0.4094 & 0.4063 & 0.4257 \\
OSULeaf & 0.0208 & 0.02 & 0.0352 & 0.0246 & 0.0463 & 0.0421 & 0.0327 & 0.0126 & 0.0203 & 0.2201 & 0.3353 & 0.2141 & 0.2412 & 0.2599 & 0.3544 & \textbf{0.4432} \\
Plane & 0.8598 & 0.8046 & 0.8414 & 0.8675 & 0.8736 & 0.9218 & 0.8784 & 0.9642 & \textbf{1} & 0.8678 & \textbf{1} & 0.8947 & 0.8947 & 0.9296 & 0.9615 & 0.9982 \\
Prox.phal.outl.ageGroup & 0.0635 & 0.0182 & 0.083 & 0.0726 & 0.0938 & 0.0682 & 0.0377 & 0.011 & 0.0332 & 0.4153 & 0.6813 & 0.25 & 0.5396 & 0.5581 & 0.6317 & \textbf{0.693} \\
Prox.phal.TW & 0.0082 & 0.0308 & 0.2215 & 0.1187 & 0.0809 & 0.1919 & 0.2167 & 0.1577 & 0.0107 & 0.6199 & \textbf{1} & 0.5864 & 0.3289 & 0.6539 & 0.733 & 0.8947 \\
SonyAIBORobotSurface & 0.6112 & 0.6122 & 0.6112 & 0.6278 & 0.6368 & 0.6129 & 0.5516 & \textbf{0.7107} & 0.5803 & 0.2559 & 0.5597 & 0.2773 & 0.4451 & 0.6634 & 0.6112 & 0.6096 \\
SonyAIBORobotSurfaceII & 0.5444 & 0.4802 & 0.5413 & 0.5107 & 0.5406 & 0.5619 & 0.5481 & 0.011 & 0.5903 & 0.4257 & 0.6858 & 0.2214 & 0.2327 & 0.6121 & 0.5647 & \textbf{0.702} \\
SwedishLeaf & 0.0168 & 0.0082 & 0.0934 & 0.0457 & 0.0269 & 0.0073 & 0.1277 & 0.1041 & 0.3456 & 0.6187 & 0.9186 & 0.5569 & 0.5573 & 0.6663 & 0.6106 & \textbf{0.934} \\
Symbols & 0.778 & 0.7277 & 0.7593 & 0.7174 & 0.8027 & 0.8264 & \textbf{0.9388} & 0.6366 & 0.8691 & 0.7995 & 0.8821 & 0.7421 & 0.7419 & 0.8989 & 0.894 & 0.9147 \\
ToeSegmentation1 & 0.0022 & 0.0089 & 0.2141 & 0.088 & 0.0174 & 0.0202 & 0.2712 & 0.3073 & 0.3073 & 0.0188 & 0.3351 & 0.001 & 0.001 & 0.3115 & 0.3671 & \textbf{0.461} \\
ToeSegmentation2 & 0.0863 & 0.0727 & 0.1713 & 0.1713 & 0.1625 & 0.0863 & 0.2627 & 0.0863 & 0.1519 & 0.0096 & 0.4308 & 0.0065 & 0.0118 & 0.3249 & \textbf{0.5498} & 0.4664 \\
TwoPatterns & 0.4696 & 0.3393 & 0.4351 & 0.4678 & 0.4608 & 0.4705 & 0.4419 & 0.3949 & 0.2979 & 0.0119 & 0.4911 & 0.0195 & 0.0142 & 0.4713 & 0.411 & \textbf{0.515} \\
TwoLeadECG & 0 & 0.0004 & 0.1353 & 0.1238 & 0.0829 & 0.0011 & 0.0103 & 0 & 0.0529 & 0.0036 & 0.5471 & 0.0017 & 0.001 & 0.4614 & \textbf{0.6911} & 0.5654 \\
Wafer & 0.001 & 0.001 & 0.0546 & 0.0746 & 0.0194 & 0.001 & 0 & 0.001 & 0.001 & 0.0008 & 0.0492 & 0.0165 & 0.0188 & 0.0228 & \textbf{0.2089} & 0.052 \\
Wine & 0.0031 & 0.0045 & 0.0259 & 0.0065 & 0.0096 & 0.0094 & 0.0211 & 0.0119 & 0.0171 & 0 & \textbf{0.7511} & 0.0018 & 0.1708 & 0.258 & 0.5927 & 0.6045 \\
WordsSynonyms & 0.5435 & 0.4745 & 0.5396 & 0.5623 & \textbf{0.5462} & 0.4874 & 0.4527 & 0.4154 & 0.3933 & 0.3498 & 0.4984 & 0.4134 & 0.4387 & 0.5448 & 0.3947 & 0.5112 \\
\hline
Total best acc              & 0     & 0            & 1    & 0     & 2     & 2    & 1    & 3     & 2    & 0      & 9 & 0  & 0  & 4  &  14 & 10 \\
Ours 1-to-1 Wins            & 34    & 34           & 32    & 31    & 33    & 31   & 34    & 33    & 33    & 36     & 24  & 35 & 35 & 23 & 21 &  - \\
Ours 1-to-1 Draws           & 0     & 0           & 1     & 1     & 1     & 2    & 0    & 2     & 1     & 0      & 4  & 0  & 1 & 0 & 2 & - \\
Ours 1-to-1 Losses          & 2    & 2           & 3    & 4    & 2     & 3    & 2    & 1     & 2    & 0      & 8 &  1 & 0 & 13 & 13 &  - \\
\hline

\hline
\end{tabular}}
\end{table*}

\subsubsection{Friedman test and Wilcoxon test}\label{sub:exp-friedman-test}
We follow the process described in~\cite{demvsar2006statistical} to do the Friedman test
and the Wilcoxon-signed rank test with Holm's $\alpha$ ($5$\%)~\cite{holm1979simple} for all the methods.

The Friedman test is a non-parametric statistical test to detect the differences in $36$ datasets across $15$ methods.
Our statistical significance is $p < 0.001$, which is smaller than $\alpha = 0.05$.
Thus, we reject the null hypothesis, and there is a significant difference among all $15$ methods.

\begin{figure}[tbp]
    \centering
    \includegraphics[width=\linewidth]{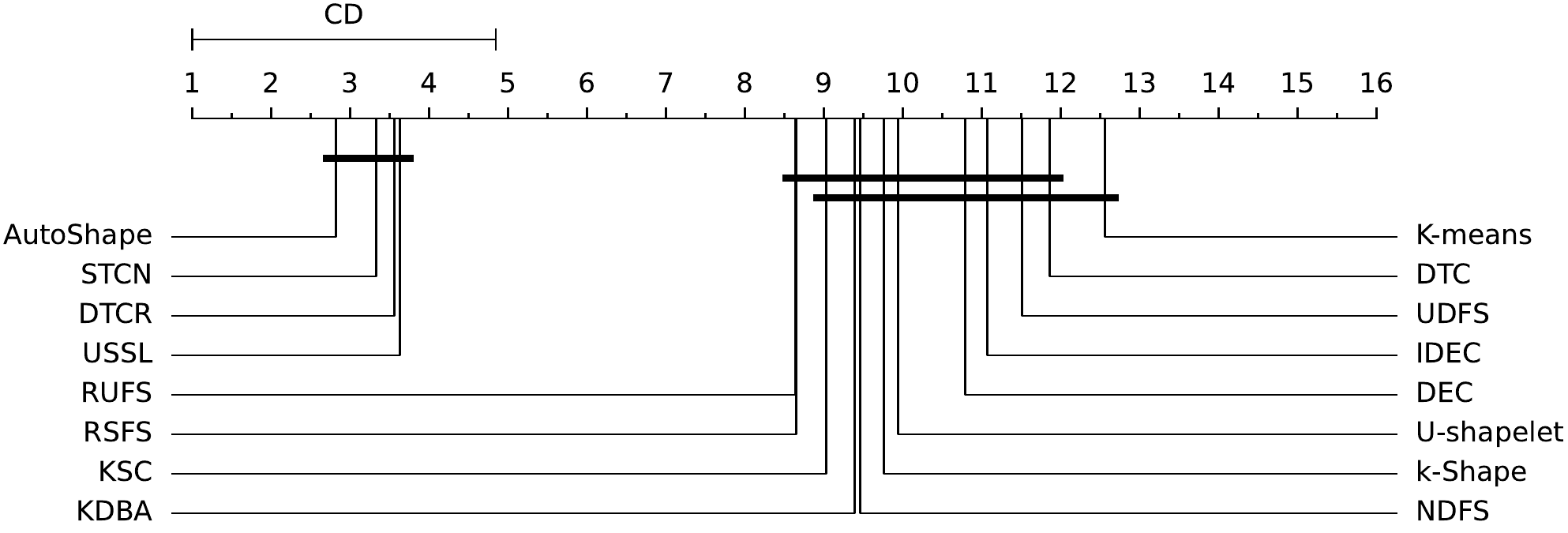}
    \caption{Critical difference diagram of the pairwise statistical comparison of $15$ methods on the $36$ datasets from UCR archive.
      \AutoShape\ ranks highest among all the compared methods.}\label{fig:cd-digram}
\end{figure}

We then conduct the post-hoc analysis among all the methods.
The results are visualized by the critical difference diagram in Figure~\ref{fig:cd-digram}.
A thick horizontal line groups a set of methods that are not significantly different.
We note that \AutoShape\ clearly outperforms other approaches except STCN, DTCR, and USSL.
However, when compared with STCN and DTCR, \AutoShape\ provides shapelets,
which are discriminative subsequences for clustering, instead of a black box.
The reconstruction loss of \AutoShape\ maintains more details of final shapelets for interpretability,
rather than learning some subsequences not in the original time series.

\subsubsection{Varying shapelet numbers}\label{sub:exp-accuracy-shapelet-number}
We compare the impact of different shapelet numbers on the final NMI of \AutoShape\ on four datasets,
BirdChicken, Coffee, SwedishLeaf, and ToeSegmentation1.

\begin{figure}[tbp]
    \centering
    \includegraphics[width=.7\linewidth]{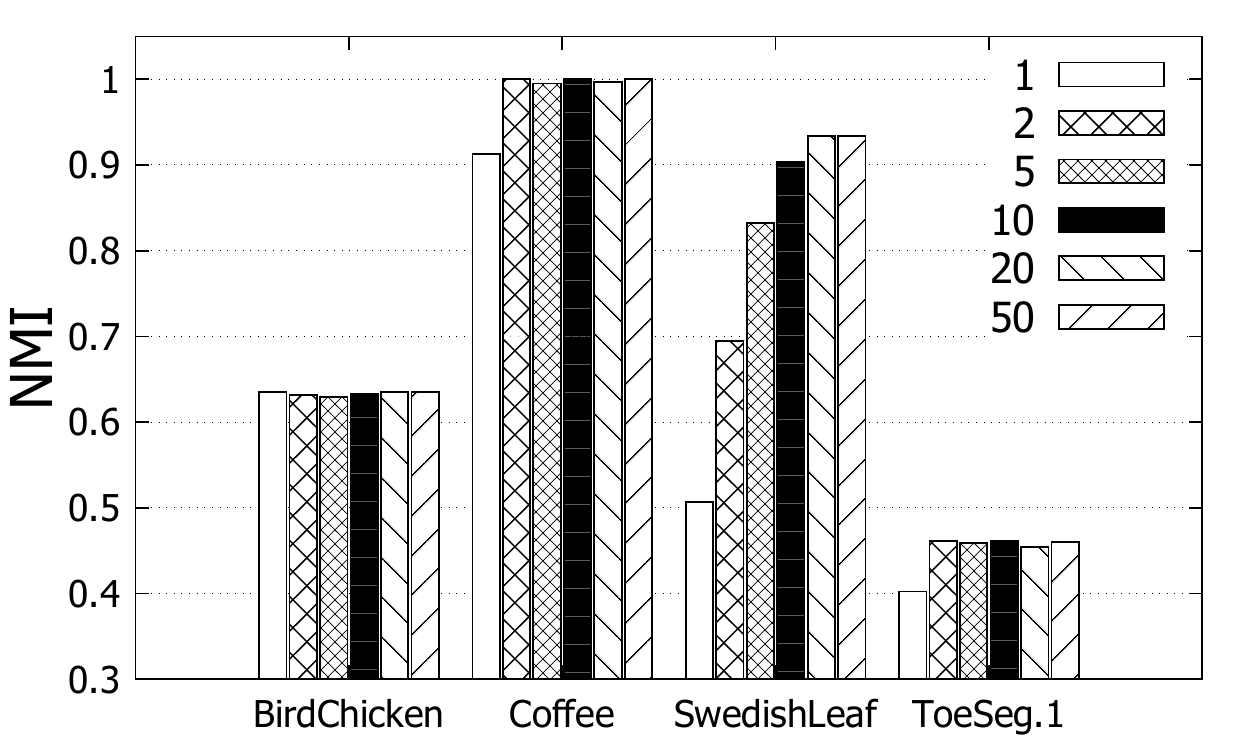}
  \caption{NMI by varying $5$ numbers of shapelet on four \UTS~datasets}\label{fig:acc-shapelet-num}
\end{figure}

Figure~\ref{fig:acc-shapelet-num} shows the NMI by varying the shapelet numbers.
Four different datasets show different trends, which guide the appropriate shapelet number of the dataset.
For example, the NMI stabilizes in BirdChicken through all the different numbers of shapelet.
Thus, the shapelet number of BirdChicken is $1$.
The NMI rises rapidly with the increase in the shapelet number from $1$ to $20$ on SwedishLeaf, and then remains stable.
Therefore, its shapelet number is set to $20$.

\subsubsection{Ablation study}
We conduct an ablation study on \AutoShape\ to further verify
the effectiveness of $\mathcal{L}_{Triplet}$, $\mathcal{L}_{Diversity}$, and $\mathcal{L}_{DBI}$.
We show a comparison between the full \AutoShape\ and its three ablation models:
\AutoShape\ without self-supervised loss (w/o triplet),
\AutoShape\ without diversity loss (w/o diversity), and \AutoShape\ without DBI loss (w/o DBI).

We observe in Table~\ref{tab:ablation} that all three elements
make important contributions to improving the final clustering performance.
The general unified representation learning (self-supervised loss) plays a particularly important role,
since the NMI result of w/o triplet is always worse than the other two losses.
In addition, we find that the shapelet candidates selection (diversity loss) and DBI objective
clearly and consistently improve the final performance.

\begin{table*}[t]
  \caption{Normalized mutual information (NMI) ablation comparisons on the $36$ datasets from {\sc UCR Archive}}\label{tab:ablation}
  \centering
  \resizebox{1\linewidth}{!}{
  \begin{tabular}{l|cccc|l|cccc}
    Dataset    & w/o triplet & w/o diversity & w/o DBI & \AutoShape & Dataset   & w/o triplet & w/o diversity & w/o DBI & \AutoShape \\
\hline
\hline

Arrow                 & 0.3942      & 0.4257        & 0.4344  & \textbf{0.5624}      & Mid.phal.outl.correct  & 0.2314      & 0.2534        & 0.2615  & \textbf{0.2873}      \\
Beef                  & 0.3345      & 0.359         & 0.367   & \textbf{0.3799}      & Mid.phal.TW            & 0.8925      & 0.926         & 0.9307  & \textbf{0.945}       \\
BeetleFly             & 0.4894      & 0.5125        & 0.5019  & \textbf{0.531}       & MoteStrain             & 0.3748      & 0.4011        & 0.3924  & \textbf{0.4257}      \\
BirdChicken           & 0.541       & 0.6124        & 0.6099  & \textbf{0.6352}      & OSULeaf                & 0.3847      & 0.4127        & 0.4009  & \textbf{0.4432}      \\
Car                   & 0.352       & 0.4347        & 0.4438  & \textbf{0.497}       & Plane                  & 0.9045      & 0.985         & 0.972   & \textbf{0.9982}      \\
ChlorineConcentration & 0.0095      & 0.0107        & 0.0103  & \textbf{0.0133}      & Prox.phal.outl.ageGroup& 0.6051      & 0.6567        & 0.6438  & \textbf{0.693}       \\
Coffee                & 0.954       & 0.9854        & 0.9799  & \textbf{1}           & Prox.phal.TW           & 0.799       & 0.883         & 0.867   & \textbf{0.8947}      \\
DiatomsizeReduction   & 0.9158      & 0.9867        & 0.9943  & \textbf{1}           & SonyAIBORobotSurface   & 0.5048      & 0.5568        & 0.5607  & \textbf{0.6096}      \\
Dist.phal.outl.agegroup & 0.375     & 0.4069        & 0.4168  & \textbf{0.44}        & SonyAIBORobotSurfaceII & 0.6287      & 0.6566        & 0.6601  & \textbf{0.702}       \\
Dist.phal.outl.correct& 0.099       & 0.1154        & 0.1106  & \textbf{0.1333}      & SwedishLeaf            & 0.843       & 0.9247        & 0.9055  & \textbf{0.934}       \\
ECG200                & 0.3048      & 0.3457        & 0.3541  & \textbf{0.3928}      & Symbols                & 0.8253      & 0.8851        & 0.893   & \textbf{0.9147}      \\
ECGFiveDays           & 0.714       & 0.756         & 0.769   & \textbf{0.7835}      & ToeSegmentation1       & 0.416       & 0.431         & 0.4301  & \textbf{0.461}       \\
GunPoint              & 0.3201      & 0.3652        & 0.3549  & \textbf{0.4027}      & ToeSegmentation2       & 0.4057      & 0.4342        & 0.449   & \textbf{0.4664}      \\
Ham                   & 0.2563      & 0.294         & 0.305   & \textbf{0.3211}      & TwoPatterns            & 0.4765      & 0.5017        & 0.5102  & \textbf{0.515}       \\
Herring               & 0.1347      & 0.1625        & 0.1599  & \textbf{0.2019}      & TwoLeadECG             & 0.4994      & 0.534         & 0.5276  & \textbf{0.5654}      \\
Lighting2             & 0.308       & 0.3241        & 0.3197  & \textbf{0.353}       & Wafer                  & 0.0257      & 0.0407        & 0.0398  & \textbf{0.052}       \\
Meat                  & 0.882       & 0.9277        & 0.9189  & \textbf{0.9437}      & Wine                   & 0.562       & 0.596         & 0.587   & \textbf{0.6045}      \\
Mid.phal.outl.agegroup& 0.321       & 0.3467        & 0.3577  & \textbf{0.394}       & WordsSynonyms          & 0.468       & 0.497         & 0.502   & \textbf{0.5112}      \\

\hline

\hline
  \end{tabular}
  }
\end{table*}

\subsubsection{Comparison with other methods on RI}\label{sub:exp-overall-accuracy-ri}
All the RI results of the baselines are taken from the original
papers~\cite{zhang2018salient} and~\cite{ma2019learning}, respectively.
The overall RI results for the $36$ UCR datasets are presented in Table~\ref{tab:ri}.

From Table~\ref{tab:ri}, we can observe that the overall performance of \AutoShape\ is ranked $1st$ among all $15$ compared methods.
Moreover, \AutoShape\ performs the best in $9$ datasets, which is more than the other methods except STCN.
\csgzli{The 1-to-1 Wins RI number of \AutoShape\ is clearly larger than 1-to-1 Losses of all other methods.}
The total highest RI number of \AutoShape\ is also larger than that of USSL and DTCR except STCN, and clearly more than those of other methods.
\AutoShape\ achieves clearly much higher RI number in some datasets, such as BirdChicken and ToeSegmentation1.
Our results on 1-to-1-Losses datasets are only slightly lower than those of USSL (\eg~Meat, SonyAIBORobotSurface) and DTCR (\eg~Lighting2, Wine).

\begin{center}
\begin{table*}
  \caption{Rand Index (RI) comparison on the $36$ datasets from {\sc UCR Archive}}\label{tab:ri}
  \centering
  \resizebox{\linewidth}{!}{
    \begin{tabular}{l|rrrrrrrrrrrrrrrr}
      Dataset & K-means & UDFS   & NDFS   & RUFS   & RSFS & KSC & KDBA & k-Shape & U-shapelet & DTC & USSL & DEC & IDEC & DTCR & STCN  & Ours   \\
      \hline
      \hline

Arrow & 0.6905 & 0.7254 & 0.7381 & \textbf{0.7476} & 0.7108 & 0.7254 & 0.7222 & 0.7254 & 0.646 & 0.6692 & 0.7159 & 0.5817 & 0.621 & 0.6868 & 0.7234 & 0.7342 \\
Beef & 0.6713 & 0.6759 & 0.7034 & 0.7149 & 0.6975 & 0.7057 & 0.6713 & 0.5402 & 0.6966 & 0.6345 & 0.6966 & 0.5954 & 0.6276 & \textbf{0.8046} & 0.7471 & 0.7784 \\
BeetleFly & 0.4789 & 0.4949 & 0.5579 & 0.6053 & 0.6516 & 0.6053 & 0.6052 & 0.6053 & 0.7314 & 0.5211 & 0.8105 & 0.4947 & 0.6053 & 0.9 & \textbf{1} & 0.8827 \\
BirdChicken & 0.4947 & 0.4947 & 0.7316 & 0.5579 & 0.6632 & 0.7316 & 0.6053 & 0.6632 & 0.5579 & 0.4947 & 0.8105 & 0.4737 & 0.4789 & 0.8105 & \textbf{1} & 0.8345 \\
Car & 0.6345 & 0.6757 & 0.626 & 0.6667 & 0.6708 & 0.6898 & 0.6254 & 0.7028 & 0.6418 & 0.6695 & 0.7345 & 0.6859 & 0.687 & 0.7501 & 0.7372 & \textbf{0.7743} \\
ChlorineConcentration & 0.5241 & 0.5282 & 0.5225 & 0.533 & 0.5316 & 0.5256 & 0.53 & 0.4111 & 0.5318 & 0.5353 & 0.4997 & 0.5348 & 0.535 & 0.5357 & \textbf{0.5384} & 0.5291 \\
Coffee & 0.746 & 0.8624 & \textbf{1} & 0.5476 & \textbf{1} & \textbf{1} & 0.4851 & \textbf{1} & \textbf{1} & 0.4841 & \textbf{1} & 0.4921 & 0.5767 & 0.9286 & \textbf{1} & \textbf{1} \\
DiatomsizeReduction & 0.9583 & 0.9583 & 0.9583 & 0.9333 & 0.9137 & \textbf{1} & 0.9583 & \textbf{1} & 0.7083 & 0.8792 & \textbf{1} & 0.9294 & 0.7347 & 0.9682 & 0.9921 & \textbf{1} \\
Dist.phal.outl.agegroup & 0.6171 & 0.6531 & 0.6239 & 0.6252 & 0.6539 & 0.6535 & 0.675 & 0.602 & 0.6273 & 0.7812 & 0.665 & 0.7785 & 0.7786 & 0.7825 & \textbf{0.7825} & 0.7647 \\
Dist.phal.outl.correct & 0.5252 & 0.5362 & 0.5362 & 0.5252 & 0.5327 & 0.5235 & 0.5203 & 0.5252 & 0.5098 & 0.501 & 0.5962 & 0.5029 & 0.533 & 0.6075 & \textbf{0.6277} & 0.6258 \\
ECG200 & 0.6315 & 0.6533 & 0.6315 & 0.7018 & 0.6916 & 0.6315 & 0.6018 & 0.7018 & 0.5758 & 0.6018 & 0.7285 & 0.6422 & 0.6233 & 0.6648 & 0.7081 & \textbf{0.7586} \\
ECGFiveDays & 0.4783 & 0.502 & 0.5573 & 0.502 & 0.5953 & 0.5257 & 0.5573 & 0.502 & 0.5968 & 0.5016 & 0.834 & 0.5103 & 0.5114 & \textbf{0.9638} & 0.6504 & 0.8863 \\
GunPoint & 0.4971 & 0.5029 & 0.5102 & 0.6498 & 0.4994 & 0.4971 & 0.542 & 0.6278 & 0.6278 & 0.54 & 0.7257 & 0.4981 & 0.4974 & 0.6398 & \textbf{0.7575} & 0.7049 \\
Ham & 0.5025 & 0.5219 & 0.5362 & 0.5107 & 0.5127 & 0.5362 & 0.5141 & 0.5311 & 0.5362 & 0.5648 & \textbf{0.6393} & 0.5963 & 0.4956 & 0.5362 & 0.5879 & 0.6543 \\
Herring & 0.4965 & 0.5099 & 0.5164 & 0.5238 & 0.5151 & 0.494 & 0.5164 & 0.4965 & 0.5417 & 0.5045 & 0.619 & 0.5099 & 0.5099 & 0.5759 & 0.6036 & \textbf{0.6286} \\
Lighting2 & 0.4966 & 0.5119 & 0.5373 & 0.5729 & 0.5269 & 0.6263 & 0.5119 & 0.6548 & 0.5192 & 0.577 & 0.6955 & 0.5311 & 0.5519 & 0.5913 & 0.6786 & \textbf{0.7205} \\
Meat & 0.6595 & 0.6483 & 0.6635 & 0.6578 & 0.6657 & 0.6723 & 0.6816 & 0.6575 & 0.6742 & 0.322 & 0.774 & 0.6475 & 0.622 & \textbf{0.9763} & 0.9186 & 0.9257 \\
Mid.phal.outl.agegroup & 0.5351 & 0.5269 & 0.535 & 0.5315 & 0.5473 & 0.5364 & 0.5513 & 0.5105 & 0.5396 & 0.5757 & 0.5807 & 0.7059 & 0.68 & \textbf{0.7982} & 0.7975 & 0.7538 \\
Mid.phal.outl.correct & 0.5 & 0.5431 & 0.5047 & 0.5114 & 0.5149 & 0.5014 & 0.5563 & 0.5114 & 0.5218 & 0.5272 & \textbf{0.6635} & 0.5423 & 0.5423 & 0.5617 & 0.5442 & 0.6524 \\
Mid.phal.TW & 0.0983 & 0.1225 & 0.1919 & 0.792 & 0.8062 & 0.8187 & 0.8046 & 0.6213 & 0.792 & 0.7115 & 0.792 & 0.859 & 0.8626 & \textbf{0.8638} & 0.8625 & 0.7724 \\
MoteStrain & 0.4947 & 0.5579 & 0.6053 & 0.5579 & 0.6168 & 0.6632 & 0.4789 & 0.6053 & 0.4789 & 0.5062 & \textbf{0.8105} & 0.7435 & 0.7324 & 0.7686 & 0.656 & 0.7958 \\
OSULeaf & 0.5615 & 0.5372 & 0.5622 & 0.5497 & 0.5665 & 0.5714 & 0.5541 & 0.5538 & 0.5525 & 0.7329 & 0.6551 & 0.7484 & 0.7607 & \textbf{0.7739} & 0.7615 & 0.6425 \\
Plane & 0.9081 & 0.8949 & 0.8954 & 0.922 & 0.9314 & 0.9603 & 0.9225 & 0.9901 & \textbf{1} & 0.904 & \textbf{1} & 0.9447 & 0.9447 & 0.9549 & 0.9663 & \textbf{1} \\
Prox.phal.outl.ageGroup & 0.5288 & 0.4997 & 0.5463 & 0.578 & 0.5384 & 0.5305 & 0.5192 & 0.5617 & 0.5206 & 0.743 & 0.7939 & 0.4263 & 0.8091 & 0.8091 & \textbf{0.8379} & 0.7849 \\
Prox.phal.TW & 0.4789 & 0.4947 & 0.6053 & 0.5579 & 0.5211 & 0.6053 & 0.5211 & 0.5211 & 0.4789 & 0.838 & 0.7282 & 0.8189 & \textbf{0.903} & 0.9023 & 0.8984 & 0.7049 \\
SonyAIBORobotSurface & 0.7721 & 0.7695 & 0.7721 & 0.7787 & 0.7928 & 0.7726 & 0.7988 & 0.8084 & 0.7639 & 0.5563 & 0.8105 & 0.5732 & 0.69 & 0.8769 & 0.7356 & \textbf{0.8954} \\
SonyAIBORobotSurfaceII & 0.8697 & 0.8745 & 0.8865 & 0.8756 & 0.8948 & \textbf{0.9039} & 0.8684 & 0.5617 & 0.877 & 0.7012 & 0.8575 & 0.6514 & 0.6572 & 0.8354 & 0.7417 & 0.8987 \\
SwedishLeaf & 0.4987 & 0.4923 & 0.55 & 0.5192 & 0.5038 & 0.4923 & 0.55 & 0.5333 & 0.6154 & 0.8871 & 0.8547 & 0.8837 & 0.8893 & \textbf{0.9223} & 0.8872 & 0.9018 \\
Symbols & 0.881 & 0.8548 & 0.8562 & 0.8525 & 0.906 & 0.8982 & \textbf{0.9774} & 0.8373 & 0.9603 & 0.9053 & 0.92 & 0.8841 & 0.8857 & 0.9168 & 0.9088 & 0.9119 \\
ToeSegmentation1 & 0.4873 & 0.4921 & 0.5873 & 0.5429 & 0.4968 & 0.5 & 0.6143 & 0.6143 & 0.5873 & 0.5077 & 0.6718 & 0.4984 & 0.5017 & 0.5659 & \textbf{0.8177} & 0.7041 \\
ToeSegmentation2 & 0.5257 & 0.5257 & 0.5968 & 0.5968 & 0.5826 & 0.5257 & 0.5573 & 0.5257 & 0.502 & 0.5348 & 0.6778 & 0.4991 & 0.4991 & 0.8286 & 0.8186 & \textbf{0.8546} \\
TwoPatterns & 0.8529 & 0.8259 & 0.853 & 0.8385 & 0.8588 & \textbf{0.8585} & 0.8446 & 0.8046 & 0.7757 & 0.6251 & 0.8318 & 0.6293 & 0.6338 & 0.6984 & 0.7619 & 0.8217 \\
TwoLeadECG & 0.5476 & 0.5495 & 0.6328 & 0.8246 & 0.5635 & 0.5464 & 0.5476 & 0.8246 & 0.5404 & 0.5116 & 0.8628 & 0.5007 & 0.5016 & 0.7114 & \textbf{0.9486} & 0.8556 \\
Wafer & 0.4925 & 0.4925 & 0.5263 & 0.5263 & 0.4925 & 0.4925 & 0.4925 & 0.4925 & 0.4925 & 0.5324 & 0.8246 & 0.5679 & 0.5597 & 0.7338 & \textbf{0.8433} & 0.8094 \\
Wine & 0.4984 & 0.4987 & 0.5123 & 0.5021 & 0.5033 & 0.5006 & 0.5064 & 0.5001 & 0.5033 & 0.4906 & \textbf{0.8985} & 0.4913 & 0.5157 & 0.6271 & 0.6925 & 0.8814 \\
WordsSynonyms & 0.8775 & 0.8697 & 0.876 & 0.8861 & 0.8817 & 0.8727 & 0.8159 & 0.7844 & 0.823 & 0.8855 & 0.854 & 0.8893 & 0.8947 & \textbf{0.8984} & 0.8748 & 0.8641 \\
\hline
Rank                        & 12.5 & 11.53       & 8.97   & 9.11   & 8.63  & 8.82  & 9.67   & 10.06 & 9.99  & 10.65  & 4.72 & 10.43  & 9.26 & 4.17  &3.94 & 3.56 \\
Total best acc              & 0     & 0           & 1      & 1      & 1     & 4    & 1      & 2     & 2     & 0      & 7    & 0      & 1    & 8     & 11  &  9 \\
Ours 1-to-1 Wins            & 34    & 34          & 32     & 31     & 31    & 30   & 32     & 31    & 33    & 32     & 20    & 31    & 30   & 23    & 20  &  - \\
Ours 1-to-1 Draws           & 0     & 0           & 1      & 0      & 1     & 2    & 0      & 2     & 1     & 0      & 3     & 0     & 0    & 0     &  1  &  - \\
Ours 1-to-1 Losses          & 2     & 2           & 3      & 5      & 4     & 4    & 4      & 3     & 2     & 4      & 13    & 5     & 6    & 13    & 15  &  - \\
Wilcoxon P-value            & 0     & 0           & 0      & 0      & 0     & 0    & 0      & 0     & 0     & 0      & 0.002 & 0     & 0    & 0.049 &  0.612  &  - \\
\hline

\hline
\end{tabular}}
\end{table*}
\end{center}

\subsubsection{Network comparison}\label{network-comparision}
We compare the performance of using Temporal Convolutional Network (TCN)~\cite{bai2018empirical}
and a recurrent network (\eg~vanilla RNN~\cite{yoon2019time}) for an autoencoder.

\begin{figure}[tbp]
  \centering
  \subfigure[Comparison on NMI]{
    \includegraphics[width=.22\textwidth]{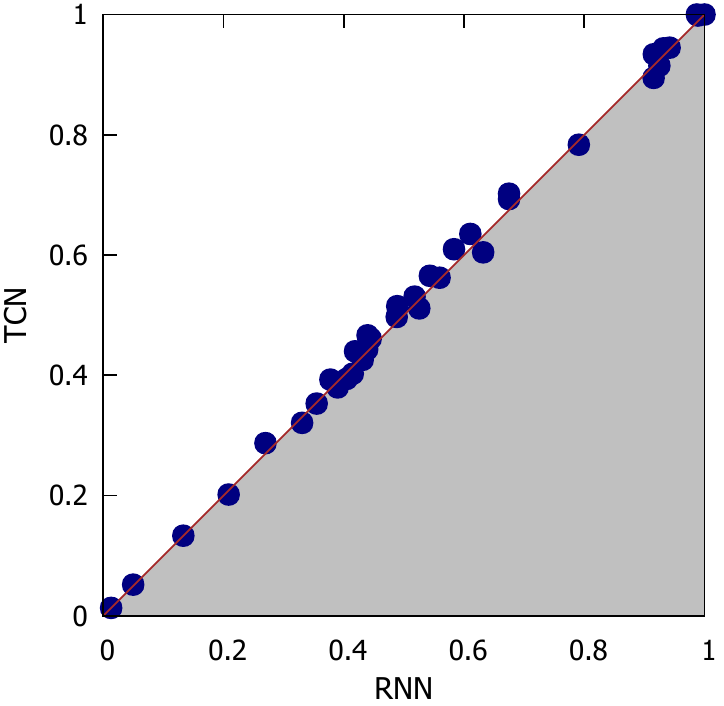}\label{fig:net-nmi}
  }
  \subfigure[Comparison on RI]{
    \includegraphics[width=.22\textwidth]{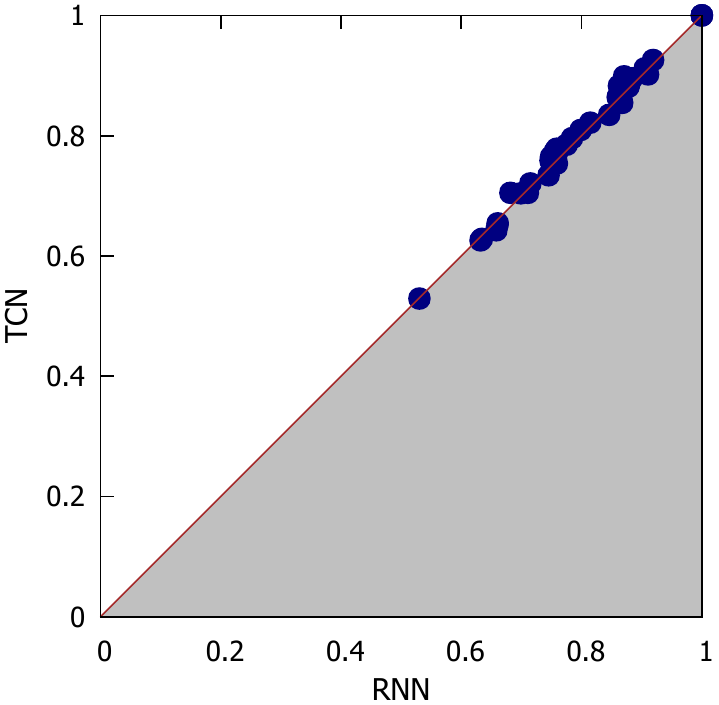}\label{fig:net-ri}
  }
  \caption{(a) The comparison of TCN and vanilla RNN over 36 datasets on NMI;
    and (b) The comparison of TCN and vanilla RNN over 36 datasets on RI.
    The dots above the diagonal indicate datasets over which TCN has better performance than vanilla RNN; vanilla RNN performs better, otherwise.}
\end{figure}

We show a comparison between the TCN and vanilla RNN on the final NMI (in Figure~\ref{fig:net-nmi}) and RI (in Figure~\ref{fig:net-ri}) performance.
The final result shows that the differences in NMI and RI between TCN and vanilla RNN are negligible in most datasets.
The statistical test reveals no evidence that either network is better than the other.
The network obey causal ordering (\ie~no future value impacts the current value) is the only requirement.

\subsubsection{Experiments on \UTS~interpretability}\label{sub:exp-interpretability}
\begin{figure}[tbp]
  \centering
  \includegraphics[width=1\linewidth]{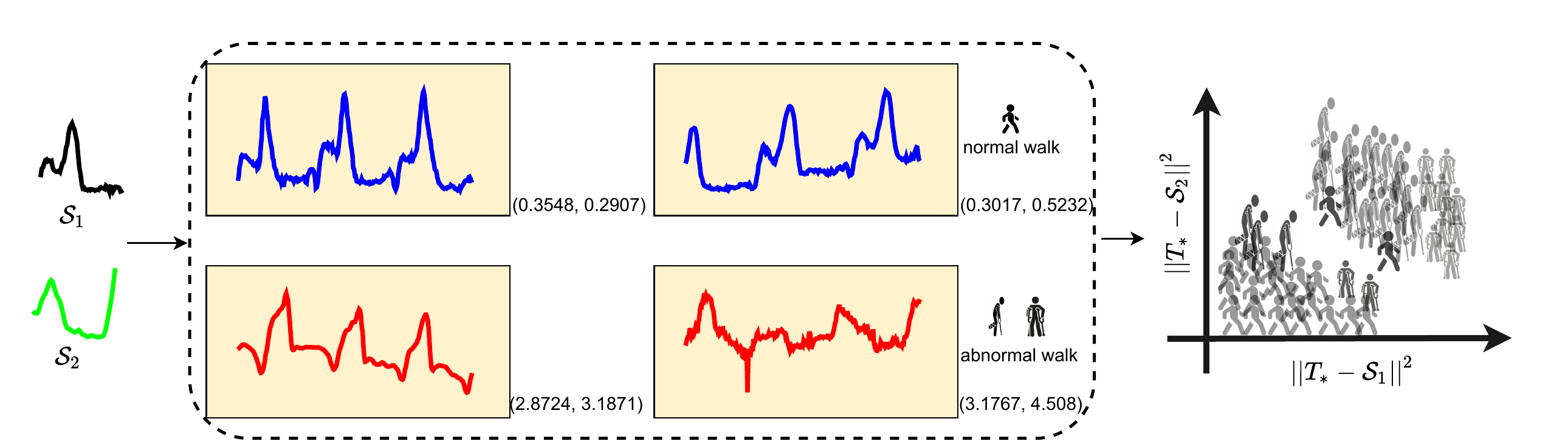}
  \caption{The top-2 shapelets $S_1$ and $S_2$ (leftmost plots) are discovered from the ToeSegmentation1 dataset by \AutoShape.
    The middle plots show four time series instances from two classes of the dataset.
    Different colors show different classes (blue for normal walk and red for abnormal walk).
    The distances to the shapelets and the original time series are used to project them into a 2-dimensional space (rightmost plot).}\label{fig:toe1}
\end{figure}

\begin{figure}[tbp]
    \centering
    \includegraphics[width=1\linewidth]{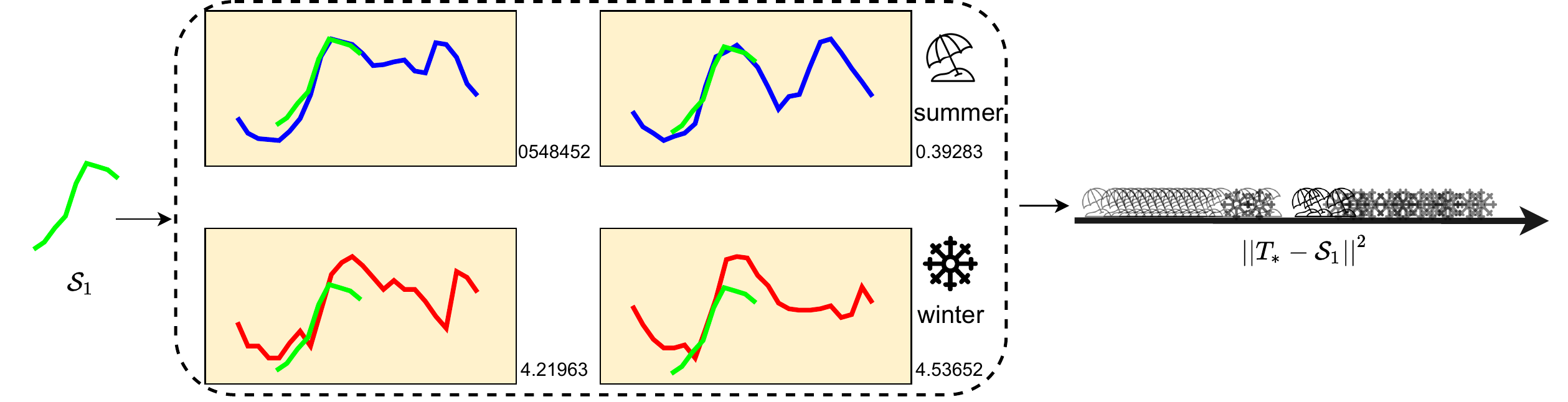}
    \caption{An example of shapelet transformation on ItalyPowerDemand.
    The original time series from summer class (blue) and winter class (red) are displayed in the middle.
    We can cluster the transformed representations into two classes effectively.}\label{fig:italy}
\end{figure}

We further investigate the interpretability of the shapelets,
which is a strength of shapelet-based methods.
We report the shapelets ($k$ =1, 2) generated by \AutoShape\ from two datasets.
These datasets are chosen simply because they can be presented without much domain knowledge.
From Figures~\ref{fig:toe1} and \ref{fig:italy},
we can observe that some subsequences of the original time series of the clusters are similar to their shapelets,
which is the reason why they are clustered.

\stab\noindent
{\bf Case study 1: ToeSegmentation1.}\label{sub:interpretability-toe1}
The ToeSegmentation1 dataset~\cite{UCRArchive}
is the left toe of z-axis value of human gait recognition (first used in~\cite{ye2011time})
derived from CMU Graphics Lab Motion Capture Database(CMU).
The dataset consists of two categories, ``normal walk'' and ``abnormal walk''
containing hobble walk, hurt leg walk and others.
In the abnormal walk category, the actors are pretending to have difficulty walking normally.

From Figure~\ref{fig:toe1}, we can easily find that the top-2 shapelets $S_1$ and $S_2$ are more frequent in normal walk class.
$S_1$  shows one unit of normal walk and $S_2$ presents the interval of two consecutive walk units.

\stab\noindent
{\bf Case study 2: ItalyPowerDemand.}\label{sub:interpretability-italy}
ItalyPowerDemand~\cite{UCRArchive}
was derived from $12$ monthly electrical power consumption time series from Italy in 1997.
Two classes are in the dataset, summer from April to September, winter from October to March.

$S_1$ is learned by \AutoShape~showed in the left part of Figure~\ref{fig:italy}.
From the learned shapelet, we can identify that the power demand in summer is lower than that in winter from 5am to 11pm.
This is because the heating in the morning used during the winter time
and air conditioning in summer was still fairly rare in Italy when the data were collected.

\subsection{Experiments on multivariate time series}\label{exp-mts}
Next, we highlight the major results from the experiments conducted on \MTS.
We take NMI~\cite{zhang2018salient} as the metric for evaluating the methods on \MTS.
We omit the rand index (RI) results, as they exhibit similar trends.
We choose K-means, GMM, k-Shape, USSL and DTCR as the compared methods
\footnote{Although we only obtained the source codes of these five methods to conduct the clustering experiments on the UEA \MTS~datasets,
  the state-of-the-art methods (\eg USSL, DTCR) are included.}.
\csgzli{Since they do not consider \MTS~datasets,
we simply concatenate the different variables into one variable for the \MTS~version (\eg k-Shape-M, USSL-M, and DTCR-M).}
The results of all $6$ methods are the mean values of $10$ runs
and the standard deviations of all the datasets are less than $0.01$.

\subsubsection{NMI on multivariate time series}
The overall NMI results for the 30 UEA \MTS~datasets are presented in Table~\ref{tab:nmi-mts}.

From Table~\ref{tab:nmi-mts}, we can observe that
the overall performance of \AutoShape\ ranks first among all the $6$ compared methods.
Moreover, \AutoShape\ performs the best in $22$ \MTS~datasets,
which is significantly more than the other $5$ methods.
The results show that \AutoShape~can learn the shapelets of high quality from different variables.

\begin{table}[t]
  \caption{Normalized mutual information (NMI) comparison on the {\sc UEA Archive}}\label{tab:nmi-mts}
  \centering
  \resizebox{\linewidth}{!}{
  \begin{tabular}{l|cc|ccc|c}
    Dataset & K-means & GMM & k-Shape-M & USSL-M & DTCR-M & \AutoShape \\
\hline
\hline

    ArticularyWordRecognition & 0.91 & 0.92 & 0.90 & 0.40 & 0.83 & \textbf{0.93} \\
    AtrialFibrillation & 0.00 & 0.00 & 0.33 & 0.44 & 0.23 & \textbf{0.50} \\
    BasicMotions & 0.05 & 0.00 & \textbf{0.94} & 0.40 & 0.50 & 0.55 \\
    CharacterTrajectories & 0.72 & 0.51 & \textbf{0.79} & 0.50 & 0.52 & 0.71 \\
    Cricket & 0.80 & 0.88 & \textbf{0.96} & 0.79 & 0.84 & 0.85 \\
    DuckDuckGeese & 0.04 & 0.04 & 0.25 & 0.31 & 0.32 & \textbf{0.35} \\
    EigenWorms & 0.05 & 0.04 & 0.22 & 0.35 & 0.36 & \textbf{0.42} \\
    Epilepsy & 0.30 & 0.25 & 0.37 & 0.40 & 0.20 & \textbf{0.42} \\
    ERing & 0.35 & 0.33 & 0.30 & 0.31 & 0.36 & \textbf{0.37} \\
    EthanolConcentration & 0.02 & 0.01 & 0.01 & 0.10 & 0.05 & \textbf{0.14} \\
    FaceDetection & 0.00 & 0.00 & 0.00 & 0.06 & 0.05 & \textbf{0.18} \\
    FingerMovements & 0.02 & 0.00 & 0.01 & 0.02 & 0.01 & \textbf{0.04} \\
    HandMovementDirection & 0.02 & 0.04 & 0.10 & 0.10 & 0.08 & \textbf{0.18} \\
    Handwriting & 0.24 & 0.23 & 0.48 & 0.44 & 0.46 & \textbf{0.51} \\
    Heartbeat & 0.01 & 0.00 & 0.04 & 0.05 & 0.07 & \textbf{0.10} \\
    InsectWingbeat & 0.01 & 0.03 & 0.06 & 0.09 & \textbf{0.10} & \textbf{0.10} \\
    JapaneseVowels & 0.72 & 0.63 & \textbf{0.75} & 0.67 & 0.47 & 0.72 \\
    Libras & 0.62 & 0.49 & 0.62 & 0.61 & 0.63 & \textbf{0.66} \\
    LSST & 0.03 & 0.00 & 0.22 & 0.20 & 0.26 & \textbf{0.28} \\
    MotorImagery & 0.01 & 0.01 & 0.01 & 0.02 & 0.08 & \textbf{0.10} \\
    NATOPS & 0.57 & 0.55 & 0.57 & 0.55 & 0.28 & \textbf{0.58} \\
    PEMS-SF & 0.33 & 0.36 & 0.36 & 0.41 & \textbf{0.46} & 0.43 \\
    PenDigits & 0.66 & \textbf{0.75} & 0.17 & 0.68 & 0.47 & 0.70 \\
    Phoneme & 0.15 & 0.08 & 0.17 & 0.21 & \textbf{0.24} & \textbf{0.24} \\
    RacketSports & 0.37 & 0.15 & \textbf{0.51} & \textbf{0.51} & 0.27 & 0.46 \\
    SelfRegulationSCP1 & 0.12 & 0.00 & 0.18 & 0.55 & 0.11 & \textbf{0.58} \\
    SelfRegulationSCP2 & 0.00 & 0.00 & 0.001 & 0.58 & 0.00 & \textbf{0.60} \\
    SpokenArabicDigits & 0.70 & 0.17 & \textbf{0.88} & 0.29 & 0.22 & 0.45 \\
    StandWalkJump & 0.00 & 0.00 & 0.18 & 0.51 & 0.23 & \textbf{0.52} \\
    UWaveGestureLibrary & 0.68 & 0.67 & 0.55 & 0.69 & 0.25 & \textbf{0.71} \\
\hline
Total best acc              & 0     & 1         & 6     & 1     & 3     & 22 \\
Ours 1-to-1 Wins            & 27    & 29        & 24    & 29    & 27    &  - \\
Ours 1-to-1 Draws           & 1     & 0         & 0     & 0     & 1     &  - \\
Ours 1-to-1 Losses          & 2     & 1         & 6     & 1     & 2     &  - \\
\hline

\hline
  \end{tabular}
  }
\end{table}

\subsubsection{Friedman test and wilcoxon test}\label{sub:exp-friedman-test-mts}
Again, we follow the process described in~\cite{demvsar2006statistical} to do the Friedman test
and the Wilcoxon-signed rank test with Holm's $\alpha$ ($5$\%)~\cite{holm1979simple}.

The Friedman test is used to detect the differences in $30$ UEA datasets across $6$ methods.
\csgzli{Our statistical significance is $p < 0.001$, which is smaller than $\alpha = 0.05$.}
Thus, we reject the null hypothesis, and there is a significant difference among all $6$ methods.

\begin{figure}[tbp]
    \centering
    \includegraphics[width=\linewidth]{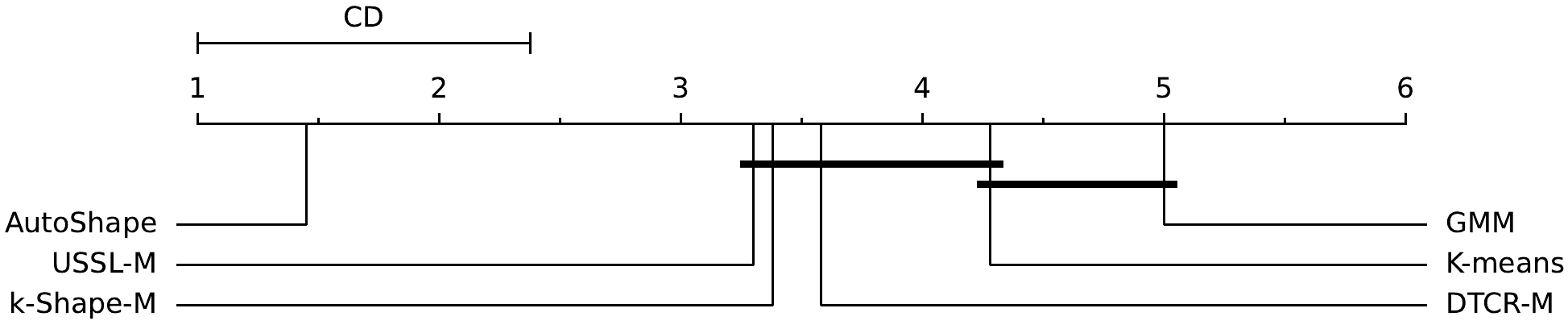}
    \caption{Critical difference diagram of the pairwise statistical comparison of $6$ methods on the UEA archive.
      \AutoShape\ outperforms all other $5$ methods significantly}\label{fig:cd-digram-mts}
\end{figure}

We then conduct the post-hoc analysis among all the compared methods.
The results are visualized by the critical difference diagram in Figure~\ref{fig:cd-digram-mts}.
We can clearly note that \AutoShape\ outperforms all other $5$ methods significantly.

\subsubsection{Varying shapelet numbers}\label{sub:exp-accuracy-shapelet-number-mts}
We further compare the impact of different shapelet numbers on the final NMI of \AutoShape\ on four \MTS~datasets,
BasicMotions, Epilepsy, SelfRegulationSCP1, and StandWalkJump.

\begin{figure}[tbp]
    \centering
    \includegraphics[width=.7\linewidth]{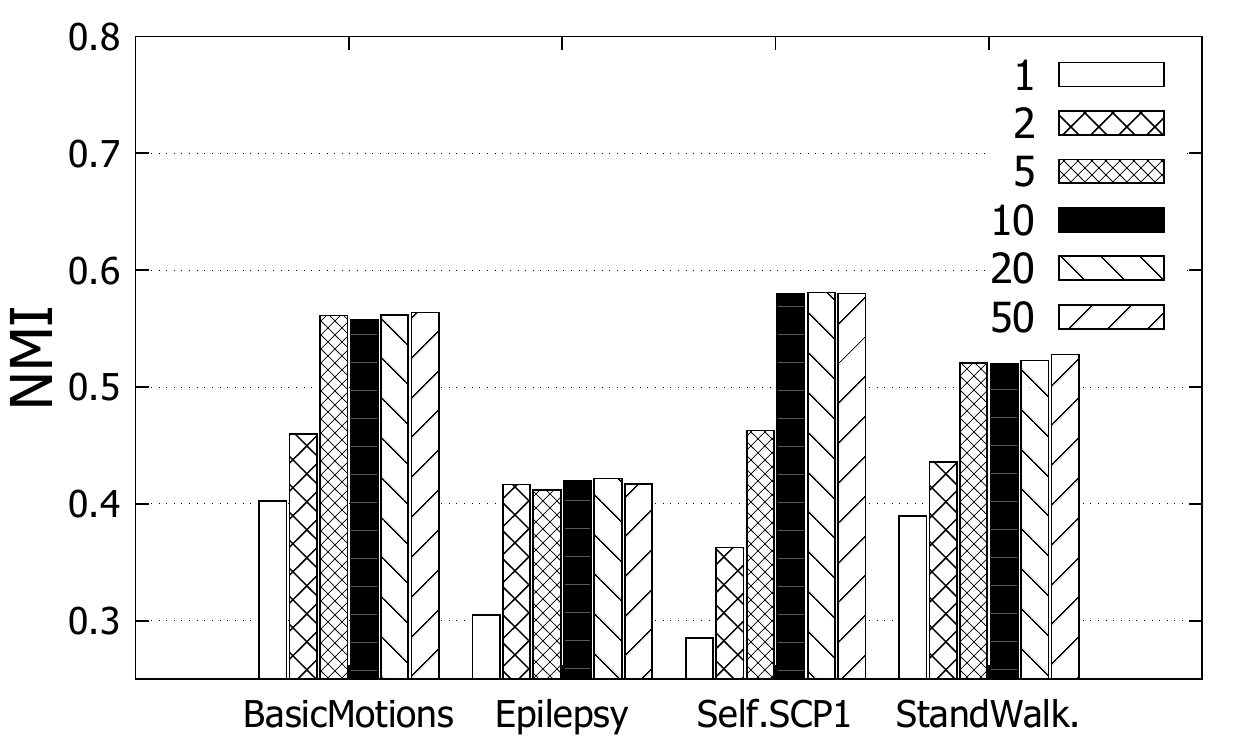}
  \caption{NMI by varying $5$ numbers of shapelet on four \MTS~datasets}\label{fig:acc-shapelet-num-mts}
\end{figure}

Figure~\ref{fig:acc-shapelet-num-mts} presents NMI by varying the $6$ different shapelet numbers.
Four different datasets exhibit different trends, which show the appropriate shapelet number selection for the dataset.
For example, NMI increases rapidly with the increase in the shapelet number from $1$ to $2$ on Epilepsy, and then remains stable.
Thus, its shapelet number is set to $2$.

\subsubsection{Experiments on \MTS~interpretability}\label{sub:exp-interpretability-mts}
\begin{figure}[tbp]
  \centering
  \includegraphics[width=\linewidth]{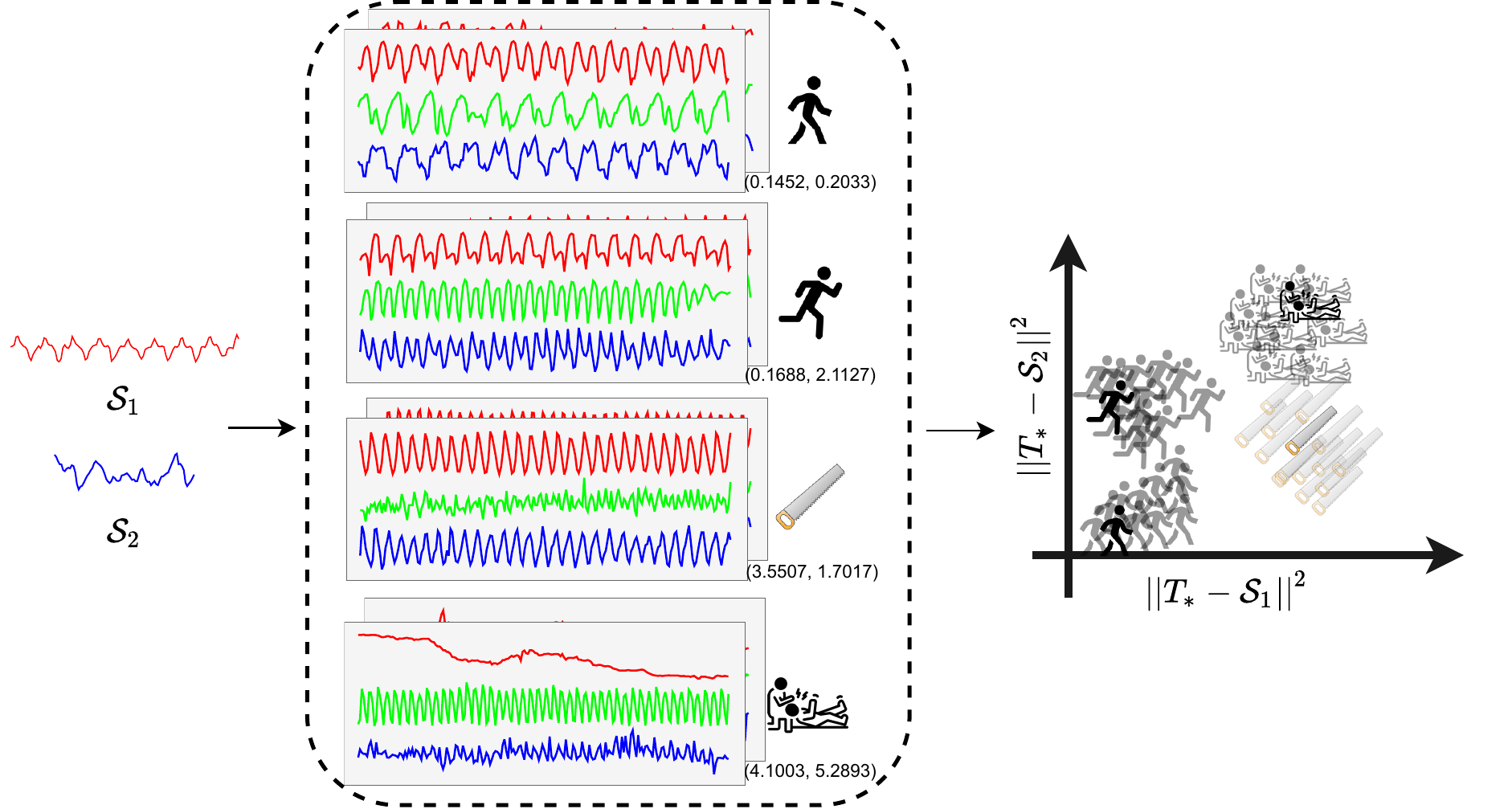}
  \caption{The top-2 shapelets $S_1$ and $S_2$ are learned from the Epilepsy dataset (leftmost plots) by \AutoShape.
    The middle plots show the time series instances from four classes.
    Three different colors show the different variables.
    The original time series are projected into a 2-dimensional space (rightmost plot).
    Most of the original time series can be correctly clustered.}\label{fig:mts-inter-epil}
\end{figure}

Finally, we investigate the interpretability of the learned shapelets on \MTS.
We describe the shapelets (\eg~$k$ = 2) generated by \AutoShape\
from the Epilepsy dataset~\cite{UEAArchive} in Figures~\ref{fig:mts-inter-epil}.
Again, we choose the dataset simply because it can be illustrated without much domain knowledge.
The Epilepsy dataset was generated with healthy participants simulating the class activities performed.
The dataset consists of $4$ categories, ``walking'', ``running'', ``sawing'' and ``seizure mimicking''.

%% file: conclusion.tex

\section{Conclusion}\label{conclusion}
This paper has proposed a novel autoencoder-based shapelet approach for time series clustering, called \AutoShape.
We propose an autoencoder network to learn the unified embeddings of shapelet candidates via the following objectives.
Self-supervised loss is for learning the general embedding of time series subsequences (shapelet candidates).
We propose diversity loss among shapelet candidates to select diverse candidates.
The reconstruction loss maintains the original time series' shapes for interpretability.
DBI is an internal index to guide network learning for improving clustering performance.
Extensive experiments show the superiority of our \AutoShape\
to the other $14$ and $5$ compared methods on \UTS~and \MTS~datasets, respectively.
The interpretability of the learned shapelets is illustrated
with three case studies on the UCR \UTS~datasets and one case study on the UEA \MTS~datasets.
As for the future work, we plan to study the efficiency and the missing values in shapelet-based methods for \TSC.

%% file: bio.tex

\begin{IEEEbiography}[{\includegraphics[width=1.1in,height=1.3in,clip,keepaspectratio]{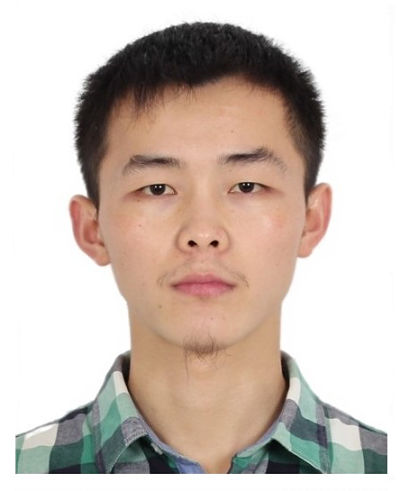}}]{Guozhong~Li}
  is a Post-doctoral Research Fellow in the Department of Computer Science, Hong Kong Baptist University.
  He received the Ph.D. degree in computer science from Hong Kong Baptist University, HKSAR, China, in 2021.
  His research interests include time series analysis, neural networking, and graph mining.
  He is a member of the Database Group at Hong Kong Baptist University (http://www.comp.hkbu.edu.hk/$\sim$db/).
\end{IEEEbiography}

\begin{IEEEbiography}[{\includegraphics[width=1.1in,height=1.3in,clip,keepaspectratio]{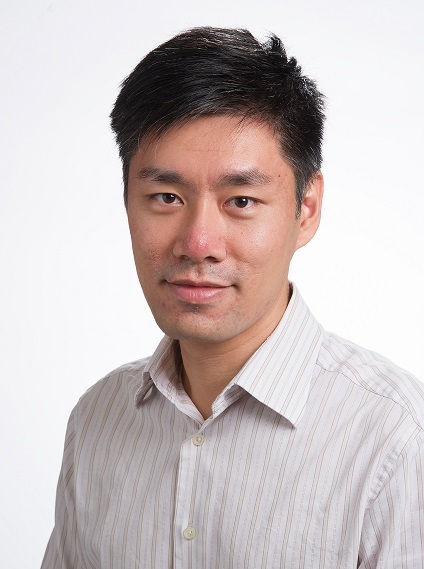}}]{Byron~Choi}
  is an Associate Professor in the Department of Computer Science at the Hong Kong Baptist University.
  He received the bachelor of engineering degree in computer engineering from the Hong Kong University of Science and Technology
(HKUST) in 1999 and the MSE and PhD degrees in computer and information science
from the University of Pennsylvania in 2002 and 2006, respectively.
\end{IEEEbiography}

\begin{IEEEbiography}[{\includegraphics[width=1.1in,height=1.3in,clip,keepaspectratio]{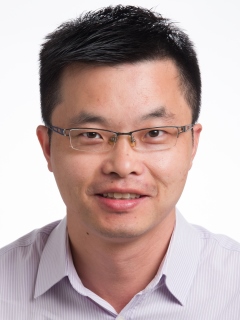}}]{Jianliang~Xu}
is a Professor in the Department of Computer Science, Hong Kong
Baptist University (HKBU). He held visiting positions at Pennsylvania State
University and Fudan University.
He has published more than 150 technical papers in these areas, most of which appeared
in leading journals and conferences including SIGMOD, VLDB, ICDE, TODS, TKDE, and VLDBJ.
\end{IEEEbiography}

\begin{IEEEbiography}[{\includegraphics[width=1.1in,height=1.3in,clip,keepaspectratio]{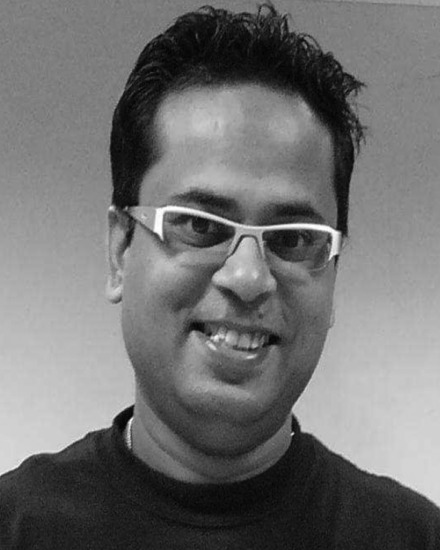}}]{Sourav~S~Bhowmick}
is an Associate Professor in the School of Computer Science
and Engineering, Nanyang Technological University. Sourav's current research
interests include data management, data analytics, computational social science,
and computational systems biology. He has published many papers in major venues
in these areas such as SIGMOD, VLDB, ICDE, SIGKDD, MM, TKDE, VLDB Journal, and
Bioinformatics.
\end{IEEEbiography}

\begin{IEEEbiography}[{\includegraphics[width=1.1in,height=1.3in,clip,keepaspectratio]{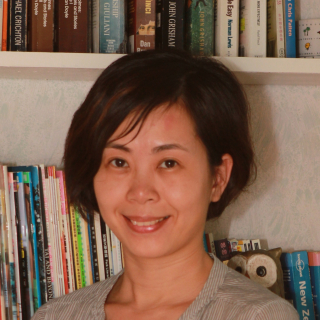}}]{Daphne~Ngar-yin~Mah}
  is Director of Asian Energy Studies Centre, and Associate Professor at Department of Geography at Hong Kong Baptist University.
  Her research focuses on social aspects of smart energy transitions, specialising in interdisciplinary research
  that cuts across the fields of energy technologies (smart grids, solar power, wind energy, nuclear power, and building energy efficiency),
  energy governance, and sustainability policy studies, with a geographical focus on East Asia covering empirical cases in China, South Korea, and Japan.
\end{IEEEbiography}

\begin{IEEEbiography}[{\includegraphics[width=1.1in,height=1.3in,clip,keepaspectratio]{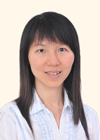}}]{Grace~L.H.Wong}
  is a Professor in the Department of Medicine \& Therapeutics, Consultant Hepatologist, Center for Liver Health, The Chinese University of Hong Kong.
  Dr. Grace Wong has published over 160 articles in peer-reviewed journals including Gastroenterology, Hepatology and Gut.
  She is currently a reviewer of 44 biomedical journals, the editorial board members of 5 journals and associate editor of Journal of Gastroenterology and Hepatology.
\end{IEEEbiography}

\vfill